\documentclass[journal]{IEEEtran}
\usepackage{lscape}
\usepackage{array}
\usepackage{graphicx}
\usepackage{multicol}
\usepackage{multirow}
\usepackage{amssymb}
\usepackage{pifont}
\usepackage{tikz}
\usepackage{verbatim}
\usepackage{array}
\usepackage{longtable}
\usepackage{supertabular}
\usepackage{amsmath}
\usepackage{epstopdf}
\usepackage{verse}
\usepackage{cite}
\usepackage{color}
\usepackage{lettrine}
\usepackage{hyperref}
\usepackage{graphicx}
\usepackage{soul}
\usepackage{comment}
\usepackage{verbatim}
\usepackage{authblk}
\usepackage{balance}
\usepackage[sort,numbers]{natbib}
\usepackage{tikz}

\usepackage{amsmath}
\usepackage{pifont}

\newcolumntype{a}{>{\columncolor{LightCyan}}c}
\begin{document}

\title{Explainable AI over the Internet of Things (IoT): Overview, State-of-the-Art and Future Directions}
\author{Senthil~Kumar Jagatheesaperumal,
        Quoc-Viet~Pham,~\IEEEmembership{Member,~IEEE},
        Rukhsana Ruby,~\IEEEmembership{Member,~IEEE},
        Zhaohui~Yang,~\IEEEmembership{Member,~IEEE}, 
        Chunmei~Xu, and
        Zhaoyang Zhang,~\IEEEmembership{Senior Member,~IEEE}

\thanks{Senthil Kumar Jagatheesaperumal is with Department of Electronics and Communication Engineering, Mepco Schlenk Engineering College, Sivakasi, Tamil Nadu, India (e-mail: senthilkumarj@mepcoeng.ac.in).}

\thanks{Quoc-Viet Pham is with the Korean Southeast Center for the 4th Industrial Revolution Leader Education, Pusan National University, Busan 46241, Republic of Korea (e-mail: vietpq@ieee.org). }

\thanks{Rukhsana Ruby is with College of Computer Science and Software Engineering, Shenzhen University, Shenzhen, Guangdong, China
(e-mail: ruby@szu.edu.cn).}

\thanks{Zhaohui Yang and Zhaoyang Zhang are with the College of Information Science and Electronic Engineering, Zhejiang University, Hangzhou 310027, China, and Zhejiang Provincial Key Lab of Information Processing, Communication and Networking (IPCAN), Hangzhou 310007, China. Zhaohui Yang is also with Zhejiang Lab, Hangzhou 31121, China.(e-mails: yang\_zhaohui@zju.edu.cn, ning\_ming@zju.edu.cn).}

\thanks{Chunmei Xu is with the School of Information Science and Engineering, Southeast University (e-mail: xuchunmei@seu.edu.cn).}

\thanks{The work of Senthil Kumar Jagatheesaperumal was supported in part by the Mepco Incubation Centre-Sivakasi (Reg. No. SRG/Virudhunagar/21/2020) and Romica Technologies, Sivakasi, India  (Reg. No.
FR/Virudhunagar/42/2022). The work of Quoc-Viet Pham was supported in part by the National Research Foundation of Korea (NRF) Grant funded by the Korean Government (MSIT) under Grant NRF-2019R1C1C1006143 and in part by BK21 Four, Korean Southeast Center for the 4th Industrial Revolution Leader Education. The work of Zhaohui Yang and Zhaoyang Zhang was supported in part by the National Natural Science Foundation of China under Grant U20A20158 and 61725104, National Key R\&D Program of China under Grant 2020YFB1807101 and 2018YFB1801104.}
}
 
\maketitle
\begin{abstract}
Explainable Artificial Intelligence (XAI) is transforming the field of Artificial Intelligence (AI) by enhancing the trust of end-users in machines. As the number of connected devices keeps on growing, the Internet of Things (IoT) market needs to be trustworthy for the end-users. However, existing literature still lacks a systematic and comprehensive survey work on the use of XAI for IoT. To bridge this lacking, in this paper, we address the XAI frameworks with a focus on their characteristics and support for IoT. We illustrate the widely-used XAI services for IoT applications, such as security enhancement, Internet of Medical Things (IoMT), Industrial IoT (IIoT), and Internet of City Things (IoCT). We also suggest the implementation choice of XAI models over IoT systems in these applications with appropriate examples and summarize the key inferences for future works. Moreover, we present the cutting-edge development in edge XAI structures and the support of sixth-generation (6G) communication services for IoT applications, along with key inferences. In a nutshell, this paper constitutes the first holistic compilation on the development of XAI-based frameworks tailored for the demands of future IoT use cases.
\end{abstract}
\IEEEpeerreviewmaketitle
\section{Introduction}
\IEEEPARstart{E}{Explainable} Artificial Intelligence (XAI) raises increasing attention towards various applications owing to having numerous advantages, such as highly transparent, trustworthy, and interpretable system development. 
Artificial Intelligence (AI) systems are being evolved day by day with more sophisticated features. AI has also grown to the level such that human brains can directly interface with machines. It has become a part of every business operation and day-to-day life of human beings. However, these are often prone to model bias, lack code confidence and trust issues. In order to manage such risks and keep the AI models transparent, the advent of XAI gives much meaningful interpretation of the system without any confusion on the decisions taken or on any embraced solutions~\cite{mohseni2021multidisciplinary}. Implications of XAI in the current businesses could replace the conventional AI systems, and these are capable of making a larger impact with better growth and sustainable developments in the production, manufacturing, supply chain, financial sectors, and wealth management.

Recently, XAI technology has attracted widespread interest from both industry and academia. The evolution of this technology has reached great success with trustworthy decisions made from the models. The emergence of XAI currently spans over a large range of applications that drive investments in various research domains. Most popular applications of XAI include healthcare~\cite{nazar2021systematic}, finance~\cite{kute2021deep}, security~\cite{zolanvari2021trust}, military~\cite{warren2020friend}, and legal sector~\cite{deeks2019judicial}. Generally, the XAI technology has proven its potential in a way that currently requires interpretable AI models. A practical example that employs XAI is the defense sector~\cite{gunning2019darpa}. In addition, the cloud services by Google are exploring the potential of XAI for deploying interpretable and inclusive AI models~\cite{kumara2022focloud}. 

As one of the most successful impacts of XAI for Internet of Things (IoT) environments, interpretable and transparent ML models~\cite{lipton2018mythos} promise new strategies to explain black box decision systems~\cite{guidotti2018survey}, based on the design of new explanation styles~\cite{van2021evaluating} for evaluating transparency and interpretability of AI systems. The technical aspects of algorithms used for explanation can then be used by the IoT systems to ensure ethical aspects in the XAI models. An example of the XAI system in the IoT is~\cite{zolanvari2021trust}, which has been demonstrated to facilitate the effectiveness of transparency using statistical theory for providing model-agnostic explanations in the Industrial IoT (IIoT). However, the challenge involved is the consideration of random new samples, which needs to be addressed for high-risk IoT applications. By using LIME-based XAI models, domain-invariant features can be learned to ensure trustworthy information processing, and they are capable of providing faithful explanations~\cite{walambe2022explainable}.

A pivotal challenge of XAI models is the customization of models for handling non-linear data, which can be circumvented by developing data-driven optimization of XAI models~\cite{hewitt2020data}. In particular, although completely interpretable models are in their infancy, XAI systems still require novel models that could address the theoretical and practical aspects of explanation and interpretability. For instance, privacy and data protection in IoT devices may not be handled safely enough by AI models on how the models reach their decisions. Some explanations of IoT applications have to be made obvious, particularly in the healthcare and military application that could substantially be benefited from XAI. Thus, XAI technology has the promise of becoming the trustworthy technology for IoT and its associated enabling technologies.

\subsection{Motivation}

Significant challenges imposed by the AI systems due to their opacity of black-box models often give the threat to trust from the ethical perspective~\cite{jacovi2021formalizing}. The inherent interpretable nature of XAI models is established through making decisions in a transparent manner, and it allows the sharing of the explanations without any argument. Recently, XAI models have made significant progress in terms of delivering trustworthy, transparent, and ethical decisions. For example, a large range of novel XAI models has been developed to enhance transparency in the decisions made with high ethical concerns. As a result, XAI models are also well suited for a large range of applications~\cite{gunning2019darpa}. However, XAI support of IoT applications is very limited, which has been restricted due to the resource constraints of smart devices. Extending the support of XAI to IoT applications and beyond allows academic and industrial research into a new dimension, which has the potential to uphold the ethical concerns and transparency of decisions made in healthcare, defense, industries, and other IoT-driven industrial applications.

Based on this motivation, a larger number of XAI models have been deployed in the IoT-applications, such as healthcare~\cite{nazar2021systematic}, finance~\cite{kute2021deep}, security~\cite{zolanvari2021trust}, military~\cite{warren2020friend} and legal sector~\cite{deeks2019judicial}. Furthermore, a deep XAI model for fault prediction using IoT sensors~\cite{mansouri2022deep} and an end-to-end ML model for IoT cloud systems~\cite{nguyen2021holistic} have been reported in the literature. More broadly, the novel range of XAI models can also be used to provide trustworthy explanations and could replace the conventional AI models in a large range of application domains~\cite{jagatheesaperumal2021duo}.

\subsection{Role of XAI in IoT}
With appropriate investigations of the data accumulated from different kinds of IoT devices deployed in an environment, it is feasible to perceive the activities in a particular scenario consequently. Most recognition techniques are often based on AI techniques, such as ML and DL, which could provide precise decisions. The role of AI in IoT applications could be listed under three stages of evolution as follows.

\begin{itemize}
    \item In the first category, the data are collected from sensors and IoT and then fed into an AI algorithm or an ML algorithm in the AI domain. 

    \item The second evolution is the usage of AI to improve IoT services. This can be as simple as conducting an investigation on sensor data, such as whether these are going out of bounds and trying to identify the reason for going out of bounds and whether the data should be fed into the AI domain. 

    \item The third role of the AI model is to supervise AI elements in the IoT domain and exchange information between algorithms and ML systems in the AI domain. 
\end{itemize}

With various models defined for IoT, we can observe a way for the AI domain to reason about the detail ins and outs in the IoT domain. For instance, if we intend to diagnose a fault, we can draw data from AI capabilities in the IoT domain. We can also start querying about the dependability and trustworthiness of the sensors or data sources. However, the usage of conventional AI techniques lacks in providing explanations to humans about the decisions made by the developed frameworks. Acquiring an obvious explanation of making such decisions serves the demand of multi-fold objectives for the better interpretation of the model during its development and simultaneous provision of more straightforward refined means of context-aware services. For instance, persistent checking of patient activities in healthcare applications is important for understanding the health status. More transparent monitoring of patient activities permits the specialists to completely comprehend the patient's behavior. 

XAI models integrated with meta-learning strategies are largely used in cyber-physical systems that are the core components of Industry 4.0. They ensure rich simulation infrastructure, smart communication with machines, higher level of visualization, better analysis of service quality and maximization of production efficiency~\cite{zoppi2021meta}. In another similar work presented by the authors in~\cite{zolanvari2021trust}, for imparting higher level security features in IIoT frameworks, Model-Agnostic Explanations were dealt with for addressing the cybersecurity threats in smart industries. Here, the transparency was imparted through statistical theory and provided explanations even for random new sets of samples for ensuring security in high-risk IIoT tasks. In~\cite{zolanvari2021trust}, the authors presented a trusted and evident experience platform for assessing the electrical power consumption behaviour in the IoT-enabled smart home scenarios. Further, the role of XAI in the healthcare sector is gaining importance in conjunction with IoT for disease prediction and diagnosis. The work in~\cite{srinivasu2022blackbox} deals with the XAI models that enable IoT frameworks used in the medical field to address the challenges involved in the prediction and diagnosis of diseases.

As the profound impact of XAI is largely emerging, the entire profound logic behind the decision-making phases of the AI techniques, such as ML and DL models, can be understood. Further, the XAI algorithm allows the models to interpret every individual decision during the prediction phases. Their importance in the IoT frameworks stands as a challenging means of addressing the issues involved in the implementation of XAI in resource-constrained IoT devices. However, the profound impact of XAI over IoT makes the end-users trust these devices deployed in commercial and public scenarios. With the essential ingredient of XAI models and the IoT data, the prominent issues from the perspective of end-users, on cost-effective ways of training the models and transparent decision making, could be addressed. The more models we have in practice, the more expensive these will be. If we have straightforward models, these are easy to explain, and they will be driven by the key elements of XAI that address the model accuracy, model quality, and cost-effectiveness aspects of training and deploying the models.

\subsection{Comparison and Our Contributions}

Motivated by the advancement in the fields of XAI and IoT, several related review works have been presented by the research community. Particularly, XAI frameworks have been extensively studied by various researchers over the past few years. For instance, some papers have provided a general overview of the XAI system and its features~\cite{dovsilovic2018explainable,adadi2018peeking,vilone2020explainable,arrieta2020explainable,gerlings2020reviewing,angelov2021explainable}, and a survey on different XAI algorithms~\cite{das2020opportunities}. In particular, the authors in~\cite{dovsilovic2018explainable} summarized the role of XAI in supervised learning along with its recent developments in association with artificial general intelligence. Similarly, the authors in~\cite{vilone2020explainable} review the contributions of state-of-the-art approaches in XAI with clustering strategies employed on theories, notions, methodologies and their evaluation. In~\cite{gerlings2020reviewing}, the authors identified four themes to the debate in addressing the XAI black-box problem. Further, based on the critical review, the findings contribute to enhanced knowledge of the decisions made by the XAI models. Angelov et al.~\cite{angelov2021explainable} related the advancements made in machine learning and deep learning research to the explainability concerns. Here, the authors formulated the principles of explainability and suggested future research directions in this particular field of research. In~\cite{das2020opportunities}, Das et al. proposed taxonomy and categorized the XAI techniques based on their inherent features to be configured as a self-explanatory learning model. Further, the authors evaluated eight XAI algorithms and generated explanation maps along with a summary of the limitations of the approaches. The authors in~\cite{adadi2018peeking} provided an overview of XAI, its background details extracted from AI, the origin of development, and technology standardization, along with the XAI architectures, use cases, and research challenges. Arrieta et al.~\cite{arrieta2020explainable} present a thorough survey on the taxonomy of XAI, leading to the conceptual framework of responsible AI. In addition, it also stimulates researchers to utilize AI systems with interpretable capabilities. 

With better ethical concerns, XAI provides trustworthy systems with explanations about the model. The explanations are required for improving the model, justification on the decision made by the system, controlling the system during abnormal behavior, and discovering new laws as well as hidden insights~\cite{adadi2018peeking}. In~\cite{xu2019explainable,longo2020explainable,meske2021explainable}, the authors compared various XAI frameworks from the context of deep learning (DL), automated decision making, and personalized experiences with respect to the research challenges and fields of applications. Others focus on specific functionality, such as security~\cite{mathews2019explainable,islam2021explainable}, healthcare~\cite{payrovnaziri2020explainable,tjoa2020survey,jimenez2020drug}, augmentation~\cite{vassiliades2021argumentation}, robots~\cite{anjomshoae2019explainable,sado2020explainable}, and the solutions associated with ML models~\cite{emmert2020explainable,roscher2020explainable,linardatos2021explainable} for exploring insights into the decisions made by the systems. Furthermore, the authors in~\cite{puiutta2020explainable} provided a detailed summary of using XAI in RL-based applications. Table~\ref{tab:surveys} presents some of the existing survey articles briefly from the perspective of XAI along with their key contributions and limitations.
\begin{table*}[!ht]
\centering
\caption{Existing surveys on XAI-related topics and our new contributions.} 
\label{tab:surveys}
\begin{tabular}{|p{0.45cm}|p{0.5cm}|p{2.0cm}|p{7.45cm}|p{5.5cm}|}
\hline
\textbf{Ref.} & \textbf{Year} & \textbf{Topic} & \textbf{Key contributions} & \textbf{Limitations}\\ \hline
\hline
\cite{dovsilovic2018explainable}  & 2018 & XAI concept  & A survey of recent developments in XAI from the supervised learning perspective. &  The applications of XAI from IoT perspective are not discussed. \\ \hline
\cite{adadi2018peeking}  & 2018 & XAI concept & A survey on the XAI concept with a basic introduction to
definitions, and exploring the AI black-box. & The applications of XAI in IoT networks have
not been presented.\\ \hline
\cite{xu2019explainable}  & 2019 & XAI and research challenges & A survey on the expert systems from the context of deep neural networks. & The benefits of XAI in IoT services are not focused.  \\ \hline
\cite{mathews2019explainable}  & 2019 & XAI and security & A survey of using NLP in malware classification tasks. & The paper only focuses on discussing the roles of malware classification and lacks IoT focus.  \\ \hline
\cite{anjomshoae2019explainable}  & 2019 &  XAI and robots & A systematic survey on the robots/agents explaining their actions to humans. & The paper only focuses on discussing the roles of XAI in the robotics domain.    \\ \hline
\cite{vilone2020explainable}  & 2020 & XAI concept  & A survey on the XAI concepts with focus on theories, methods and evaluation of models. &  The paper only analyzes the use of XAI in general and lacks focus on IoT domains.\\ \hline
\cite{arrieta2020explainable}  & 2020 & XAI concept & A survey on the concepts, taxonomies, opportunities and challenges in XAI-based systems. & The applications of XAI in IoT networks have not been presented.  \\ \hline
\cite{gerlings2020reviewing}  & 2020 & XAI concept & An overview on the use of XAI for addressing the black-box problem. & The benefits of XAI in IoT services are not explored.  \\ \hline
\cite{das2020opportunities}  & 2020 & XAI and its opportunities & IA summary of explanations maps of eight different XAI algorithms. &  The discussion of the XAI opportunities for IoT networks has not been provided. \\ \hline
\cite{longo2020explainable}  & 2020 & XAI and research challenges & An overview on the construction of predictive models and explanations from automated / semi-automated decision making. &  The research challenges are not concentrated on IoT frameworks. \\ \hline
\cite{fouladgar2020xai}  & 2020 & XAI practice to theory & A systematic survey on the research gap among XAI theory and practical applications. & The applications of XAI in practical IoT applications have not been explored.  \\ \hline
\cite{payrovnaziri2020explainable}  & 2020 & XAI and healthcare & An overview on the application of XAI in the medical field through electronic health records. & The applications of XAI in IoT networks and services have not been explored and discussed.  \\ \hline
\cite{tjoa2020survey}  & 2020 & XAI and healthcare & A survey on the medical XAI applications with insights to medical practitioners. & The paper only focuses on discussing the roles of XAI in the healthcare domain.  \\ \hline
\cite{jimenez2020drug}  & 2020 &  XAI and healthcare &   A systematic survey on the prominent XAI algorithms for drug discovery tasks. & Only analysis of XAI in drug discovery is provided while IoT domains are not considered. \\ \hline
\cite{vassiliades2021argumentation}  & 2021 & XAI and augmentation & A brief overview on the using augmentation for providing explanations for various applications domains. & The applications of XAI in IoT networks have not been presented.  \\ \hline
\cite{sado2020explainable}  & 2020 & XAI and robots & A survey on the explaining and communicating among robotic agents through perpetual and cognitive reasoning. & The applications of XAI in IoT networks and services have not been explored and discussed.  \\ \hline
\cite{emmert2020explainable}  & 2020 & XAI and ML & A survey on the advances of XAI and ML for technological progress in various fields. & The applications of XAI in IoT networks have not been presented.  \\ \hline
\cite{roscher2020explainable}  & 2020 & XAI and ML & A brief survey on the usage of XAI and ML for scientific insights and discoveries. & The discussion of the XAI applications for IoT networks has not been provided.  \\ \hline
\cite{puiutta2020explainable}  & 2020 & XAI and RL & A brief survey on the application of XAI in RL applications. & The paper lacks generation explanations for IoT services.  \\ \hline
\cite{stepin2021survey}  & 2021 & XAI models & An overview on the contrastive and counterfactual
explanation generation methods for XAI. & The benefits of explanation generation for IoT services are not focused.  \\ \hline
\cite{angelov2021explainable}  & 2021 & XAI concept & An analytical review on the four principles of explainability. & The paper only discusses the generic analytics of models and lacks the focus on IoT.  \\ \hline
\cite{meske2021explainable}  & 2021 & XAI and research opportunities & A systematic survey on the personalized explanations for various stakeholder groups. & The paper only focuses on generalized aspects of XAI.  \\ \hline
\cite{chou2021counterfactuals}  & 2021 & XAI and its applications & A survey of promoting causability in model-agnostic approaches for XAI. & The applications of XAI in IoT networks have not been presented.   \\ \hline
\cite{islam2021explainable}  & 2021 & XAI and security & A survey on the XAI approaches for human-centric AI systems. & The paper only focuses on discussing the roles of XAI in credit card fraud. \\ \hline
\cite{linardatos2021explainable}  & 2021 & XAI and ML & An overview on the ML interpretability methods as well their programming implementations. & Only analysis on the XAI-ML is provided while other domains are not considered.  \\ \hline
\cite{islam2022past}  & 2022 &  XAI and ML  &  A survey on the generation of contextual explanatory models & Only focuses on the contextualizing methods and no integration of IoT.  \\ \hline
\cite{ahmed2022artificial}  & 2022 &  XAI and Industry 4.0  &  A summary of the XAI models are used in Industry 4.0 applications. &  The core benefits of IoT for XAI-driven smart industries are not focused. \\ \hline
\cite{majid2022applications}  & 2022 & XAI and WSN &  A systematic survey on the solutions for smart industries through XAI and WSN are addressed. &  The benefits of XAI and WSN for IoT networks and devices are not discussed. \\ \hline
\cite{alicioglu2022survey}  & 2022 & XAI and Visual analytics &  A comprehensive survey on XAI solutions for ML models and data visualization & The benefits of XAI in IoT services are not focused.  \\ \hline
Our \newline paper & 2022 & XAI and IoT & An extensive survey on the XAI-IoT integration. Particularly, 
\begin{itemize}
    \item We discuss the roles of XAI in various key IoT services, such as security enhancement, Internet of Medical Things (IoMT), IIoT, and Internet of City Things (IoCT).
    \item The use of XAI in dependable IoT applications is also analyzed in detail, including edge XAI structures, and its potential to meet 6G requirements.
    \item A summary of key lessons learned is provided to provide insights into XAI-IoT integration. Research challenges and directions are also highlighted.
\end{itemize}  &  - \\ \hline

\end{tabular}
\end{table*}

Although XAI has been extensively studied by various researchers in literature, there does not exist a comprehensive and dedicated survey of using XAI in IoT services and applications, to the best of our knowledge. The potential of XAI in various fields, such as IoT networks, security, healthcare, and industrial sectors, has not been explored in the open literature. Moreover, a holistic summary of the integration of XAI with IoT from the perspective of smart homes to smart cities still needs to be explored. These shortcomings motivated us to conduct a comprehensive review on integrating XAI with IoT services. Specifically, we include a state-of-the-art survey of the applications of XAI in various categories of IoT applications, such as security, healthcare, industry, and smart cities. The key contributions of this article lie in the vast summary of the usage of XAI, including network security enhancement, IoMT, IIoT, and IoCT. We also summarize the key observations learned from this survey at the end of each IoT application. Finally, a discussion on the vital research challenges is provided, and promising future research directions on XAI-based IoT services are outlined. To the best of our proficiency, we are the first ones to provide a dedicated and detailed survey on the XAI in IoT systems. The main contributions of this survey can be highlighted as follows.

\begin{itemize}
    \item Compared to other related survey works in this domain, this survey provides a broad summary of relevant background details on IoT, XAI, and their integration to enable researchers to dig through the trustworthiness of IoT systems.
    
    \item We present the need for XAI in IoT and some of the key challenges presented in the recent literature and summarized some of the recent research works.
    
     \item Moreover, we explore a few of the IoT application domains, such as security, healthcare, industries, and smart cities. We present the need and role of XAI in such applications for better trustworthy explorations of the IoT services along with the lessons learnt.

    \item We also present a detailed discussion on the cutting-edge developments for dependable IoT services from the perspective of using XAI models. 

    \item Finally, we outline the future research challenges in the direction of consideration of XAI for IoT.
\end{itemize}

\subsection{Paper Structure and Organization}
This survey is organized as follows: Section~\ref{sec:XAIIoTprelims} elaborates the preliminary to XAI and IoT. Section~\ref{sec:XAIapplications} discusses the evolution of XAI systems in IoT along with the category of application in relevant domains. This section also elaborates the future scope of research in each application and the related open-end challenges. Section~\ref{sec:XAIforIoT} summarizes the key findings and outcomes of the article towards establishing the best measures of XAI-based IoT architectures.
Section V points out future directions for XAI over IoT. 
Section~\ref{sec:conclusion} concludes the paper.


\section{XAI and IoT: Preliminaries}
\label{sec:XAIIoTprelims}

In this section, we first introduce the fundamental knowledge of XAI, which is an important branch of AI. Then, we present the concepts of IoT along with its appealing characteristics. The objective of this section is to introduce the readers to the XAI, IoT, and their key principles.

\subsection{Explainable AI} 
In the year 1950, realistic confinement of the term AI was put forward by Alan Turing in his article ``Computing Machinery and Intelligence''~\cite{turing2009computing}. In this article, Alan explored the intuition of determining the thinking capabilities of machines. He models a game between a human and a computer to help identify a human by a third person who cannot see both the human and the machine involved in the game. If the third person cannot predict human competence consistently, the computer wins the game. The evolution of AI begins with the outcome of this research. The successive development of AI has grown over the years with the aid of other supporting technologies.  

With an optimized approach during the 1960s, researchers believed that AI-led expert systems would perform incredibly complex tasks. We currently reap the benefits of the research with the great success of AI in various fields, such as medical diagnosis, logistics, data mining, robotics, industrial automation, and many more. When we think of the healthcare sector, transportation, and financial decisions, AI is an all-around human being in different ways. Similar to humans who expect each other to explain their decisions, we expect the same from any AI system that can be trusted and could be understood. From the black box AI to glass box transformation, XAI enables humans to understand the system in detail. Even if we do not understand the underlying technologies, XAI systems at least enable us to understand the process. Better communication could address the different levels of trust in AI systems. The ability to explain the operational mechanism of AI and reach certain decisions is also an ultimate focus of XAI.  

\begin{figure*}
  \centering \includegraphics[width=\textwidth]{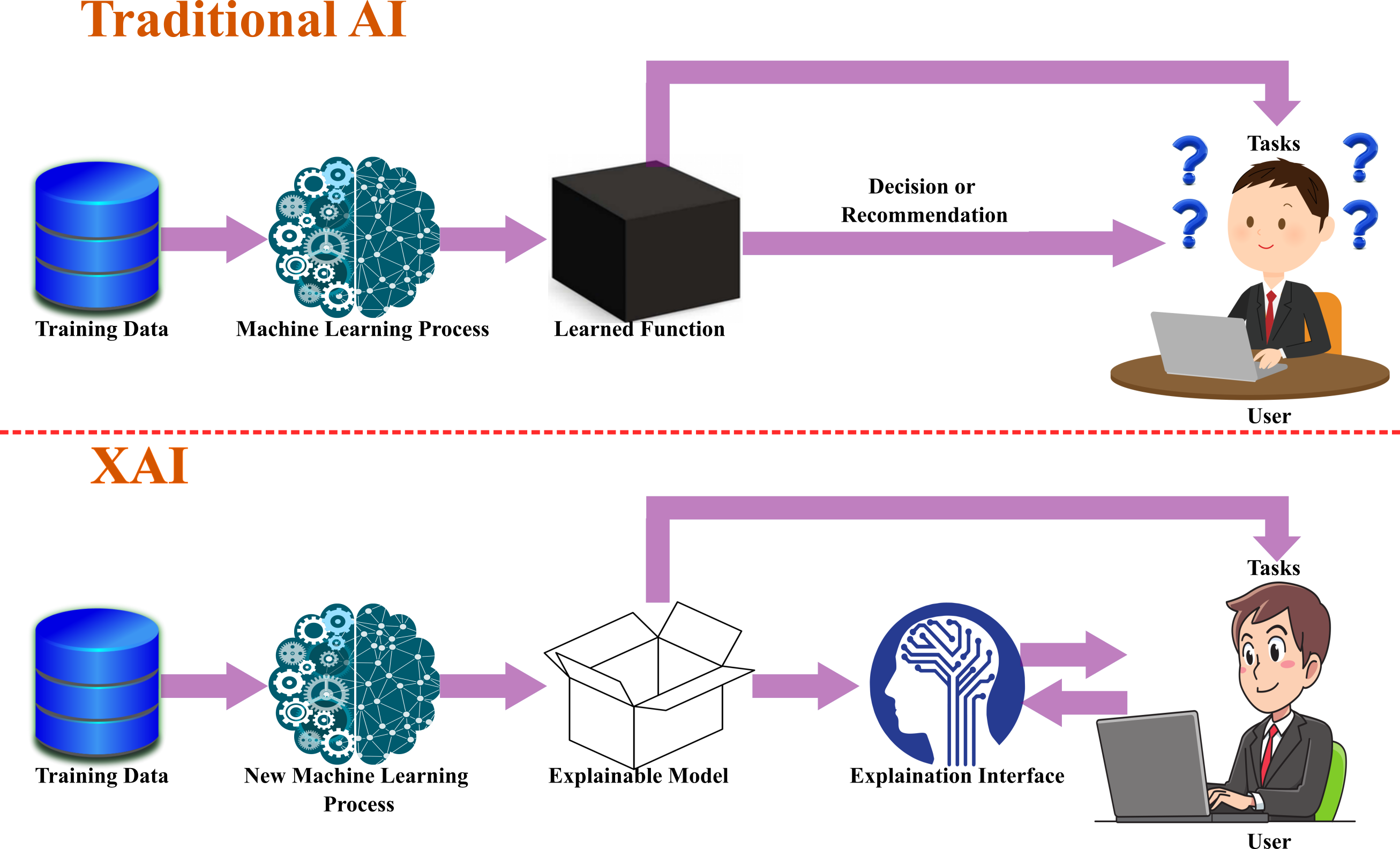}
   \caption{General sequence of operations in traditional AI and XAI.}
   \label{fig:AIXAI}
\end{figure*}

Fig.~\ref{fig:AIXAI} shows the general sequence of operations involved in traditional AI and XAI-based systems. In conventional AI systems, the learning process cannot be interpreted by the end-users, and it looks like an opaque black box. Different from conventional AI, XAI models use a revolutionary ML process with explainable models that facilitate the end-users with sustainable transparency in the learning process and the decisions made from the training data. 

Explainable model creators often require explanations of the inner workings of the model. Based on the demand from the authorized agents, by relating the output of the model to its inputs, the model examiners require explanations of the way how it has been arrived at ~\cite{meske2021explainable}. Based on the contributions of the features in the model, the explainability provides a flavor of explanations concerning the underlying data and the model. Fig.~\ref{fig:Expterms} shows the relationship between the terminologies associated with the XAI models. The detailed description of the key terminologies are elaborated as follows: 

\begin{figure}
  \centering \includegraphics[width=0.5\textwidth]{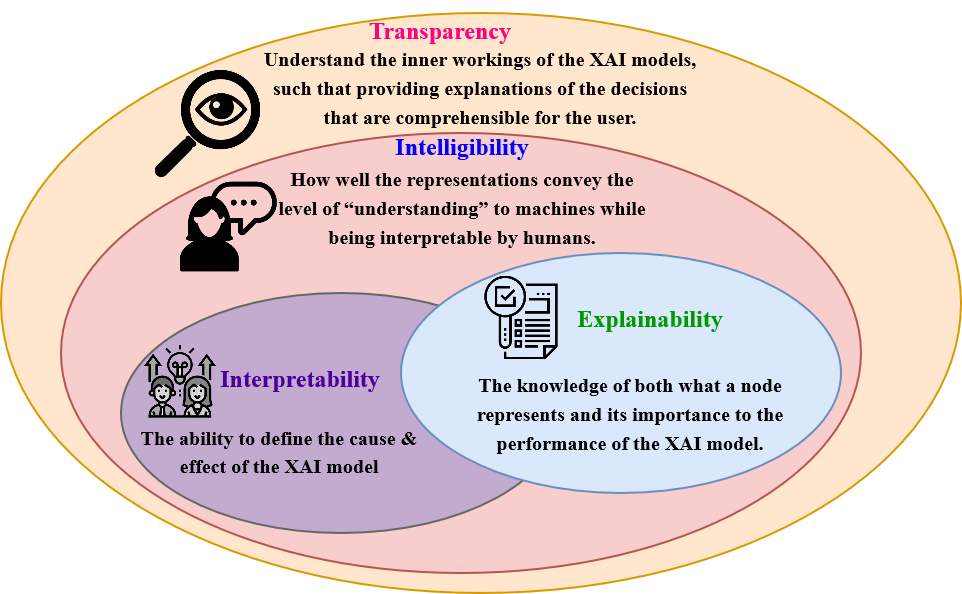}
   \caption{Relationship among the key terms associated with XAI models.}
   \label{fig:Expterms}
\end{figure}

\textit{Interpretability:} It refers to the inherent capability of the XAI models, which makes the decisions obvious and understandable to the users. Normally, the interpretability feature of the models reduces as the complexity of the model increases ~\cite{dikshit2021interpretable}. Being one of the core components of XAI, interpretability assists in managing social interactions on the decisions with clarity.

\textit{Transparency:} It is one of the crucial factors in making sound decisions, where trust is normally a challenge in AI systems. The mechanisms involved in the core functionalities of the model, including the algorithms and the data flow, could be analyzed with the inherent transparent capability of the XAI models~\cite{von2021transparency}. Conceptual analysis of the trust aspects of XAI models reveals the way how transparency becomes one of the core ingredients and, subsequently, its importance for determining the degree to which AI is fair and trustworthy. 

\textit{Trustworthiness:} In AI systems, trustworthiness refers to the models that are resilient, with their capability to protect the privacy of the users and enable secured means of interactions. Moreover, as trustworthiness is one of the key components of explainability, through confidential computing and encryption, the maximized control and security over the data are guaranteed. White-box models explain in detail the behavior, prediction process, and affecting factors. For a white-box model to be qualified as one of the robust models, the features of a model must be understood, and the learning process must be transparent~\cite{rawal2021recent}. Further, based on the type of problem at hand, it is capable of generating explanations accordingly. Moreover, white-box models have understandable features, behaviors, and exhibit strong relationships among the influencing variables and the predicted outputs.

The term transparency refers to the understanding capability of the inner working of the XAI models, such that the explanations of the decisions are comprehensive for the users. Trustworthiness, on the other hand, refers to the set of mechanisms used in the explainable layers of the XAI model that enriches the learning model in order to be trusted by the users with a clear understanding.

\begin{figure*}
  \centering \includegraphics[width=\textwidth]{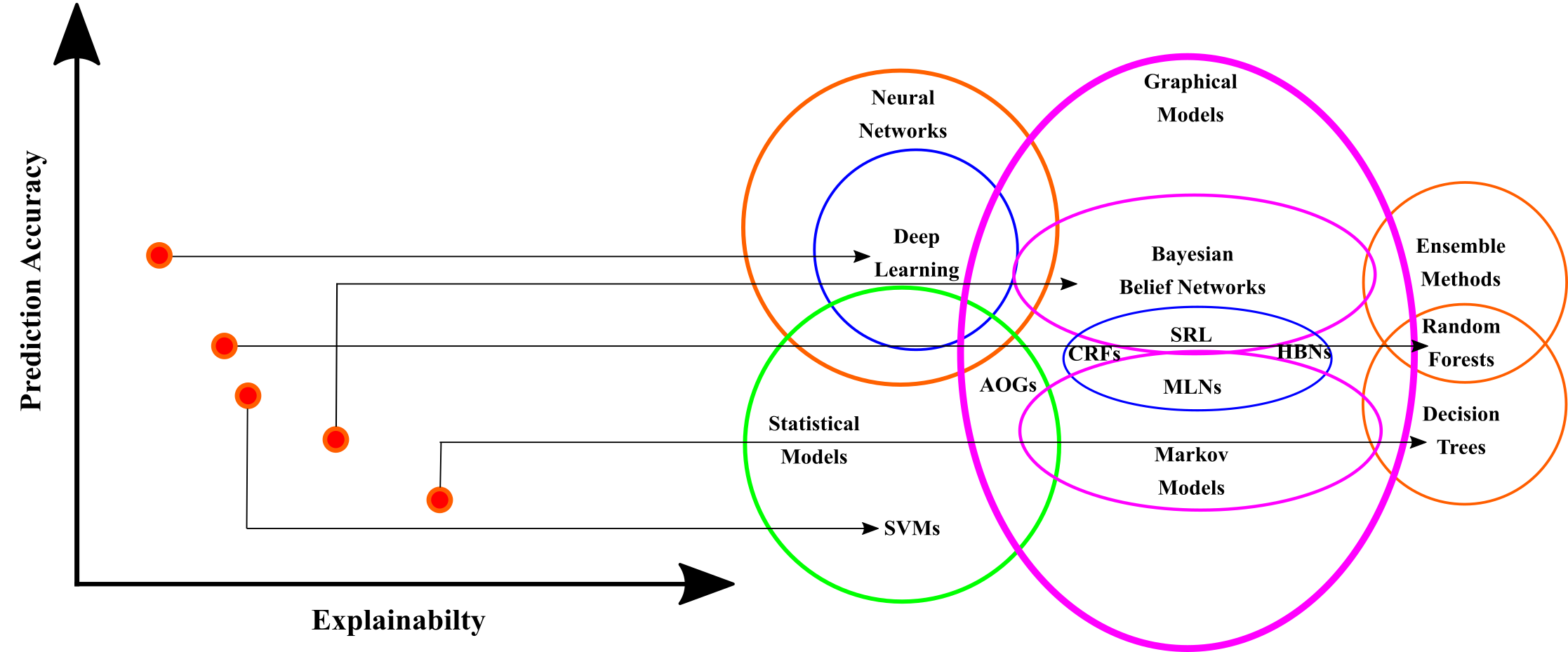}
   \caption{Accuracy prospects of XAI models based on the algorithms used for implementation.}
   \label{fig:Accuracy}
\end{figure*}

\textit{Completeness vs. Interpretability Trade-off: } In any given situation, an explainable system unfolds almost an infinite spectrum of explanations that can be provided from complete, accurate, interpretable, and easily understandable insights, on the decision taken by the model. To make the concept more concrete, a scientist working on an image classifier using the XAI concept must be capable of explaining the model so that even an ordinary person is capable of understanding the model ~\cite{van2022explainable}. Matching of explanations should be done at the levels of a developer and a user with reduced complexity. Fig.~\ref{fig:Accuracy} shows the accuracy prospects of the XAI models depending on the algorithms used for implementation. Among the popular XAI models, deep learning~\cite{nascita2021xai} models ensure better accuracy with less explainable features. However, the XAI models based on decision trees~\cite{mahbooba2021explainable}, random forests~\cite{hatwell2020chirps} and ensemble models~\cite{zou2022ensemble} provides better explainability with considerably less performance accuracy compared to deep learning models. It is also evident that there exist trade-offs on the complexity of the models, explaining capabilities and the performance of the systems.

XAI is a sub-field of AI that has been creating numerous traction, especially over the last few years. Explanations for the AI model depend on the persons who ask the questions and what their intentions are. An XAI expert tries to develop a model by exactly defining the explanations of the model with consensus. Traditional modeling techniques, such as regression or tree-based models, often have an understandable relationship between the input data and the decisions in the model outputs. For this reason, these are called white-box models. The advent of more complex modeling techniques, such as DL models and ensemble techniques, promises to provide improved model accuracy, but often at the expense of a lack of model explainability. Understanding the relationship between the input data and the output decisions is challenging in those complex modeling techniques, which form the basic building block of conventional AI-based modes. These categories of AI models are often referred to as black boxes since they may not provide explanations of the decisions taken. In some situations, it may be comfortable with the trade-offs between accuracy and explainability. For instance, we do not mind if the model recommends a particular movie as the best match, but we certainly do care about the reason for rejecting a credit card application made by someone ~\cite{kute2021deep}. It also depends on the bias in the AI models, which can be originated even before the training and testing phases. The data used for model training can be entangled in its own set of biases. Although learning the strategy to identify and handle bias in datasets is an important component of any responsible AI strategy, we require to aim to build transparent, trustworthy, and bias-free models.

The ability to understand the mechanism of making choices by a model is an important factor for three main reasons. First, it allows one to further refine and improve its analysis. Second, this becomes easier to explain to non-practitioners the way how the model uses the data to make decisions. Finally, explainability can help practitioners to avoid negative or unforeseen consequences of their models. These three factors of XAI give us more confidence in our model development and deployment.

\subsubsection{XAI Architectures for Transparency in AI}

\begin{itemize}

    \item \textit{Model-based:} Model interpretation is the core aspect of any AI-based system. It ensures to provide better decision-making policies with fairness, transparency, and accountability of the results, which may give enough confidence to humans ~\cite{yang2022design}. With the value and accuracy interpreted from the models, it can be easy to debug them and thereby carry out necessary measures to improve their performance.

    \item \textit{Probabilistic:} Probabilistic XAI models incorporate probabilistic distribution and random variables. We here define the random variable as the potential outcome of an uncertain event in the XAI model. Such uncertainties are incorporated explicitly and these are propagated through the model. The output from the probabilistic model helps to understand the uncertainty ~\cite{konforti2022sign}. This kind of forecasting is typically much more useful in providing a range of potential outcomes rather than a single best guess. Imparting explainability and trustworthy features enables the understanding of the risks involved in the forecasting process, and it helps to fine-tune the system accordingly.  
    
    \item \textit{Multi-component:} Most of the XAI system operations cannot be interpreted with single explanations and require an interactive dialogue-based system that provides multiple explanations. It enhances the social acceptance of the decisions made by the system, with a proper explanation of the decision being made. Implementing such a multi-component framework for XAI system analysis imparts deeper trust in the system ~\cite{kim2021multi}.
    
    \item \textit{Visual analytics:} With interaction representations of the decisions made by the XAI systems, the incorporation of visual analytics makes humans understand the decision-making process of AI systems with much ease. Visual analytics employs an iterative process of perceiving, interpreting, and assessing inputs and outputs of the XAI models ~\cite{andrienko2022visual}. It helps to interpret the pipeline of events triggering the actions to make an appropriate decision through the visual representation of each stage. 
    
    \item \textit{DL:} DL architectures are proven to provide strong trustworthiness for the XAI systems. Irrespective of the used resources and time taken for the interpretation of DL models, they provide a robust and most satisfactory explanation considering all the probable events in XAI systems. When the DL models with identical architecture produce distinct explanations, the accuracy of the explanations purely depends on the quality of data, size of data, and the methods used in the feature extraction process~\cite{watson2022agree}. Moreover, the set of hyperparameters used in different layers of DL models, the involved activation functions, and the carried-out optimization processes all play a significant role in defining the characteristics of any XAI system ~\cite{nascita2021xai}.

\end{itemize}

\subsubsection{XAI Frameworks for Transparency in AI}

XAI frameworks assist in producing reports about the functionality of the model and attempt to clarify their working principles with a proper explanation of the decisions made with transparency. Following are the few popular XAI frameworks that contribute towards imparting trust over the systems for users.

\begin{itemize}
    \item \textit{SHapley Additive exPlanations (SHAP):}
    This SHAP framework can provide explanations of different models, from preliminary ML algorithms to sophisticated DL and NLP approaches. The SHAP framework falls under model agnostic strategy to interpret the models, which in turn depend on Shapley values in game theory~\cite{antwarg2021explaining}. These Shapley values are computed through the average marginal contribution of a feature value over all possible combinations. The symmetry property of Shapley values explains that the features of the model which provide equal contributions obtain similar values. The dummy property of Shapley values explains that features of the model which do not provide any contributions obtain a zero value. Moreover, the additive property of the Shapley values further explains that feature contribution by adding up to the difference between prediction and average. 
    
    \item \textit{Local Interpretable Model-agnostic Explanations (LIME):}
    The LIME framework has similar features compared to SHAP, but its speed of computation is superior to SHAPE. Moreover, the LIME framework generates explanations for each sample of data to each of the predicted output results from the framework. The explanation is generated based on the assumption that every sophisticated model is linear from the perspective of a local scale~\cite{zafar2021deterministic}. It tries to fit a single observation from a simple model and mimics the behaviour of the global model in the local scenario. Then, that simple model can be configured to explain the prediction of the sophisticated model in the local scenario. 
    
    \item \textit{Explain Like I’m 5 (ELI5):} 
    It provides a simple explanation of complex XAI models, similar to a five-year-old child requesting the explanation, as that child has a limited or naive understanding of the question. From the perspective of XAI, ELI5 could be implemented to debug the ML classifiers and helps to explain their predictions~\cite{fan2019eli5}. It helps to inspect the model parameters and intercept the global characteristics of the model. It also assists in providing individual predictions of the model and valid explanations about why such decisions are made.
    
    \item \textit{What-If Tool (WIF):} 
    WIT is a visual interface designed by Google that helps to understand the data sets and outputs of the XAI models. It can be executed with minimal code in various platforms and cloud services~\cite{wexler2019if}. Analysis using the WIF tool is helpful during the creation of the model, collection of the data, and pre and post-training evaluation phases. Models developed with WIT tools can be deployed on cloud AI platforms and assist the developers in probing into their models to better understand behaviour. 
    
    \item \textit{AI Explainability 360 (AIX360):} 
    AIX360 is an open-source toolkit developed by the IBM research group to explain the arrival of AI systems at the trusted recommendation. It explains the priority of tasks involving risk factors in providing suggestions to improve the flexibility of usage of the system by end-users. Integrated explainable algorithms in the AIX360 toolkit are relevant for usage in financial sectors, and this can better be extended to other use cases as well~\cite{arya2020ai}. The algorithms included in the toolkit are highly convenient for understanding the models and data. 
    
    \item \textit{Skater:}
    Skater is an open-source tool designed to explain the learned features of the XAI model in terms of local and global prediction results~\cite{mishra2022model}. This framework provides a unified approach to enable interpretation of the models for building a transparent XAI model that can be comfortably used for most of the use cases in society. The Skater framework could use LIME and DNNs for the interpretation of local features of the data fed to the XAI model.   
    
\end{itemize}
\subsection{Internet of Things (IoT)} 

In less than a decade, there has been a great technological evolution, among which IoT is the most significant one. IoT is the interconnected web of objects and these objects are called things. There are almost $50$ billion devices that are connected to the Internet worldwide today, and in this Internet rush era, everyone wants to get into the Internet today to attain new information~\cite{chegini2021process}. IoT devices can change all the mundane tasks done by humans and make everything smart and convenient for people. These drive the technology and business sectors with smart wearable, smart buildings, smart cities, smart homes and smart everything. At the end of the day, these increase the quality of life by the click of a button on smartphones for accessing and controlling smart devices located in remote places. In IoT devices, the hardware and software work in coordination to make the exchange of information among devices easier. IoT in today's world has billions of devices connected to the web, and so we need to manage the data from the connected devices. There are zettabytes of data generated by those devices, and we can make use of these data by using data science tools to exercise our capabilities and make better decisions. Those decisions help in serving both short and long-term purposes. Therefore, in the context of IoT, the volume of data can be combined with the right software and hardware to make better decisions. Innovations in the domain of IoT envisage solutions to existing challenges and open up new possibilities for growth. In the smart healthcare sector using IoT devices, we can use the latest available hardware and software. They can be programmed and configured to probably allow the doctors to know the current health conditions of patients being in remote places and suggest probable drugs for their well-being even if they are not physically present amidst the patients. 

\subsubsection{Role of AI in IoT}

IoT is immensely useful in predicting mishaps, any possible violations, and detection and control of unexpected events in automation processes. These have enormous applications in healthcare, where physicians can gain valuable information from healthcare wearable devices, such as biochips and pacemakers. Moreover, apart from the vital data collected from healthcare devices, IoT devices employed in industrial applications, smart homes, and other smart applications generate a large volume and variety of data, which is cumbersome to manage using conventional data processing methods. To ensure accuracy and high-speed prediction and gain valuable insights from the data accumulated through IoT devices, the role of AI can make a huge impact. 

AI can be integrated with IoT devices and operate on the data collected from IoT devices for managing unplanned downtimes in industrial automation for making accurate decisions at appropriate times~\cite{lv2021ai}. AI with IoT also drives enhancing the quality and services of the products with automated inspections through the support of robotic systems. It can quickly spot defects in the products and suggest alternatives in advance. Those accurate predictions and valuable insights on the products and services also increase operational efficiency with reduced resources and manpower. 

\subsubsection{Trustworthiness of IoT Systems}

Modern-day IoT applications are largely driven by systems with the support of AI. In particular, the wireless communication architecture and protocols, along with AI models, will be able to process and transfer the data to server and cloud services for remote monitoring and controlling applications. The replacement of XAI in place of AI-based systems can spot the transparency of operations in IoT devices with more insights into the decisions taken. As heterogeneous technologies from diversified domains of engineering, science, and technology, are combined to provide satisfactory performance in IoT devices, these are highly prone to privacy and security threats. Sicari et al.~\cite{sicari2015security} survey the challenges involved in IoT security along with the requirements for enhancing the trust of the systems. Another survey by Yan et al.~\cite{yan2014survey} investigated the trust properties and provided a trust management research model for IoT from a holistic perspective. A dynamic trust management protocol~\cite{bao2012dynamic} is proposed by Bao et al. for dealing with the strange behavior of IoT systems. Furthermore, this protocol also possesses the adaptive adjustment capability based on the environment and a trust parameter. Similarly, an adaptive trust management protocol was implemented in~\cite{chen2015trust} for trustworthy social IoT services.

\section{Application Domains of XAI over IoT Use Cases}
\label{sec:XAIapplications}

In this section, we initiate with an overview of the motivation that inspires us to compile XAI solutions for IoT systems. Then, we introduce the use of XAI for various IoT applications. In particular, we discuss the mechanism of XAI in enhancing security, the healthcare sector, the Internet of Robotic Things, IIoT, and commercial IoT use cases.

\subsection{Enabling XAI in IoT systems: An Overview}

Over the last few years, XAI models have made a significant impact on the enhancement of wireless communication services and IoT~\cite{9536411}, since the evolution of IoT and smart devices has boosted the demand for imparting autonomy, as well as transparent and trustworthy interactions between human and smart IoT systems. Moreover, the progressive growth of edge computing and cloud computing platforms makes it feasible for IoT devices to store, process, and analyze the characteristics of the environment in which they are deployed~\cite{pham2020survey}. In addition, the emergence of AI further motivates the use of XAI models in enhancing the trust of data collection, processing, analyzing, and utilizing them for carrying out various autonomous tasks.

From the perspective of IoT use cases, XAI frameworks can be fit into two major locations. First, these can be used at the center of the IoT systems, which can be widely used for predictive analytics, anomaly alerts, and intelligent decision-making. Within this center stage, applications of XAI models are used to analyze the data acquired from IoT devices. For instance, XAI models can be used to analyze and predict the operational management and predictive maintenance of smart machines in industries and therefore allow the workgroup to optimize their time and resources across the manufacturing process. Moreover, the prediction of certain vital features through IoT devices can also be a prime component in smart city applications. IoT devices employed in vertical domain applications, such as smart cities, smart industry, smart vehicles, and smart grids, are capable of generating a large volume of data. Trustworthiness of the analytics platform is highly essential in such use cases. XAI models support the provision of this feature with its inherent capability of providing appropriate explanations for the decisions taken by the system, thereby providing robust smart city services. 

IoT devices are playing a crucial role in food production and agriculture. These devices installed in agricultural fields support the implementation of irrigation control systems using a network of sensors that are employed for monitoring humidity and temperature. For enhancing the interpretability and trustworthiness of automated monitoring in agricultural fields, an explainable AI-based decision support system is developed in~\cite{tsakiridis2020versatile}. The developed systems implement fuzzy rule-based automation in irrigation, which is interpretable.

Explanations are generated from IoT environments in human-centric AI domains for tracking certain aspects of users and improving support decisions, which are learned using neural networks~\cite{garcia2019human}. The developed approach is illustrated in smart IoT kitchens that detects the depression of persons based on their food intake after their meal. The results are revealed from the auto-generated explanations of human interactions with the AI system from the simulator. 

Analytics solutions for IoT devices in industries are provided through osmotic computation for solving DL problems in XAI systems~\cite{oyekanlu2018distributed}. Here, C28x DSP is used for the processing task, which saves memory usage at the edge devices in the IoT networks. 

\subsection{Security Enhancement in IoT Devices}

There are many reports on IoT devices that have been compromised by hackers due to situations where many IoT devices are being built with static credentials, weak passwords, and symmetric tokens. Patchwork security solutions and minimal built-in security features without having comprehensive security frameworks are not enough for modern IoT devices. No two IoT devices are unique, and they may have very limited computing resources and have less memory, and may not have a faster processor. Most importantly, they do not have built-in security support and may not be capable of running similar security solutions running on larger general-purpose computing systems. Most IoT devices are prone to physical attacks, such as abduction of devices, and the attackers may try to probe the device or reverse engineer a few of these to observe the corresponding activities. Attackers may also hack the firmware of the device and look for security keys and hard-coded passwords, which may lead to backdoor access to the device and other security vulnerabilities. Most IoT devices also have very long life cycles and perform critical tasks. 

\subsubsection{Security Enhancement in IoT Devices with AI}
AI and ML tools can go a long way in fighting cybercrime in IoT devices. AI-enabled analysts could respond to threats with greater confidence and speed. Even though AI could significantly improve cyber security for IoT devices, hackers may also exploit the usage of AI for criminal activity. Therefore, AI systems are widely used for both cyber defense and attacks~\cite{rahouti2021incremental}. AI can automate the authentication of IoT devices by strengthening them with strong passwords, biometric-based security features, and can protect them against cloud computing vulnerabilities. In the following, we list the most prominent advantages of deploying AI to enhance security features in IoT devices.

\begin{itemize}
    \item \textit{Handling of a large volume of data:} Data acquired from IoT devices is expanding at a steadily extending rate, and the capacity limit of big data frameworks is restricted to a limit, and hence it turns into a challenge to store and handle a large volume of information. AI frameworks driven by ML and DL can effectively handle such a large volume of data and extract meaningful predictions from them.
   
    \item \textit{Learn cybersecurity over time:} AI systems are capable of learning from a large volume of data and eventually get trained to recognize the attacks on IoT devices.
   
    \item \textit{Identifies unknown threats:} Each IoT device connected to the global network needs to be added with robust security features to avoid leakage of vital information. AI-based systems are able to spot and resist even unknown attacks on IoT data. 
    
    \item \textit{Quick detection of cybersecurity threats:} Development of high-speed computing hardware, cloud computing, and GPU have made the AI systems operate at high speed, and thereby drive them in quick detection of the threats on IoT devices.
   
    \item \textit{Saves time of human analysts:} Involving humans to detect, analyze and predict the threats in IoT devices is highly challenging. AI almost eliminates the need for human analysts, or they perform the role of supervision on the decisions taken by the AI systems. This, in turn, saves a huge level of energy and enables them to focus on other priorities. 
\end{itemize}

\subsubsection{XAI for Security Enhancement in IoT Devices}

The complication of modern attacks on the flow of data from IoT devices across networks is highly challenging to be predicted by the network administrators. Moreover, they find it cumbersome to recognize the patterns in the attacks, which may make a substantial impact on the large volume of data from IoT devices. The authors in~\cite{miloslavskaya2020stream} presented an XAI-based architecture that manages the network security-related stream of data, and they were capable of predicting the attacks against the potential network assets. Similarly, the authors in \cite{mathews2019explainable} have summed up the ongoing advancements in this field of XAI with one of the specific focuses on malware classification, with the results interpreted through appropriate visualization.

Cyber attacks against IoT systems can also be detected using Long Short-Term Memory (LSTM) based DL frameworks, which can be helpful in explaining the decision-making process~\cite{saharkhizan2020ensemble}. Here, the developed modules are integrated into an ensemble of detectors, and the robustness of the IoT devices is tested under Modbus network traffic. It was also observed that the LSTM modules provide a detection accuracy of around 99\%. In one of the similar works by Akerman et al.~\cite{akerman2019vizads}, a convolutional LSTM encoder-decoder model is adopted for detecting anomalous messages between surveillance-broadcast and air traffic control systems. More specifically, the anomalies in the image sequence are predicted, and the operative information to the pilots is explained while detecting anomalies for making relevant decisions.   

In cyber-security applications, such as Intrusion Detection Systems (IDSs), understanding the classification made by the ML techniques is possible using gradients without making changes in the model of the classifier. Further, it was observed through the experimentation performed on the NSL-KDD99 benchmark dataset that this can be extended for further diagnosis, and the interactive visualizations show the understanding and reasoning of the decision made by the classifier~\cite{marino2018adversarial}. In the context of intrusion detection carried out in~\cite{islam2019domain}, the outcome suggested that domain knowledge infusion plays a predominant role in better explainability and quicker decision-making. This feature makes the model face strange attacks and opens a way to manage huge network traffic from IoT devices. The work in \cite{amarasinghe2018improving} investigated the use of offline and online feedback mechanisms for the decision-making process of the Deep Neural Networks-based intrusion detection systems. In particular, the work focuses on qualitative analysis to enhance the transparency of the system and helps in building trust in the system.

Explainable assessment of the security capabilities for IoT applications has been investigated in~\cite{forti2020secure}, in which the authors proposed a methodology for ensuring trust relations between the cloud-edge frameworks and various stakeholders. The authors in~\cite{casola2020security} proposed an optimization process from the perspective of cloud-edge on-demand service under the implementation of security prospects. This was carried out for IIoT applications to address the cloud-edge allocation problem providing measurable enhancement in assessing the security of the system. Secure management of instant messaging services in Cloud of Things is implemented for hybrid cloud–edge environments~\cite{celesti2019approach}. Moreover, it was used for the improvement of authenticity and confidentiality in the middleware. In~\cite{shirazi2017extended}, the authors summarized the impact of extended cloud in mobile edge and fog computing systems based on resilience, anomaly detection, and security management. 

\subsubsection{Lessons Learned} 
In the aforementioned XAI-based security application of IoT devices, we have demonstrated that intrusion and attacks on IoT devices can be made transparent through XAI frameworks. To say about the key benefit, XAI assists in visualizing the impact of attacks on IoT devices to make appropriate countermeasures for establishing safe and secure data transmission among the IoT nodes. By driving the decision-making processes for modern communication systems to be obvious and understandable to stakeholders, XAI in the IoT devices supported through B5G/6G services holds responsibility for providing secured and automated actions. This, in turn, leads to a better and more robust IoT framework and thus assists in anticipating the impact of security threats in the system. The main lessons learned in this context are given as follows.

\begin{itemize}

    \item With the vast range of IoT devices that have made a paradigm shift in automating the environment monitoring and control tasks, imparting reliable and secure infrastructure is a mandated choice for those interconnected devices. Cybersecurity requirements are critical for IoT data collection, exchange of data, cloud communication, storage and analytics on the data collected from IoT devices. Further, the risks of the IoT data are also dominant despite the data in motion, or at rest, or being used by the system for processing. This scenario of interconnected devices under threats needs AI support for delivering viable and critical security services to the IoT devices in the network. As mentioned in the aforementioned sections, more reliable and transparent cyber defense decisions for mitigating the risks in the IoT devices and their networks, can be well established using XAI frameworks.
    
    \item Robust XAI models for intrusion detection and securing IoT networks were presented in this section, which can perform their predictions in a trustworthy manner and provide better explainability. In this approach, the XAI models are trained using various datasets and they can be evaluated across various feature sets. Once the data are trained, the IoT devices can then directly use the trained data for imparting trustworthy deployment and predictions. Thus, the generalisability evaluation is observed as a common feature that is particularly a vital resource for strengthening the security aspects of AI-driven IoT devices and networks. Another particular impact on the transparency is based on the complexity of the models and the volume of data used for processing. Due to these, the overhead and robustness of training the XAI models for enhancing IoT security, need to balance a trade-off among the time complexity, resources and performance accuracy.

    \item One of the major concerns in the decision-making phase on the security attacks of IoT devices depends on the challenges of the operational deployment. This is because the XAI models need to identify the source of IoT data from the devices or IoT network during the training phase, and also need to locate an appropriate space for the accumulation of data. This can further motivate the model to provide accurate classification of threats by training the diversified category of data accumulated across different IoT devices interconnected through large networks and topologies.

\end{itemize}

\subsubsection{Future Works}
If the attacks on the IoT devices are explained with a set of rules, the nature of attacks and their impact can be easily interpreted. The model or framework used for the interpretation should be able to distinguish malicious nodes from the regular flow of IoT data in the network, by analyzing the set of feature values affecting the flow of data. For instance, identification of malicious traffic is one of the key issues in dealing with the traffic of IoT networks, where XAI models can assist the network security personnel in taking a possible course of action against the malicious traffic. XAI-based IoT security enhancement can be used to determine the robust nature of IoT devices interfaced with the network by enabling the network experts to make better planning strategies in the future to avoid malicious traffic and network intrusions. Since the XAI models can be able to extract the explanations from the learning model, we can extend their capabilities towards security enhancement in IoT networks by exploiting the transparency standpoint and explainable solutions.

\subsection{Internet of Medical Things}

\subsubsection{Healthcare Support Using IoT-based Systems}
\begin{figure}[!ht]
  \centering    
  \includegraphics[width=1.0\linewidth]{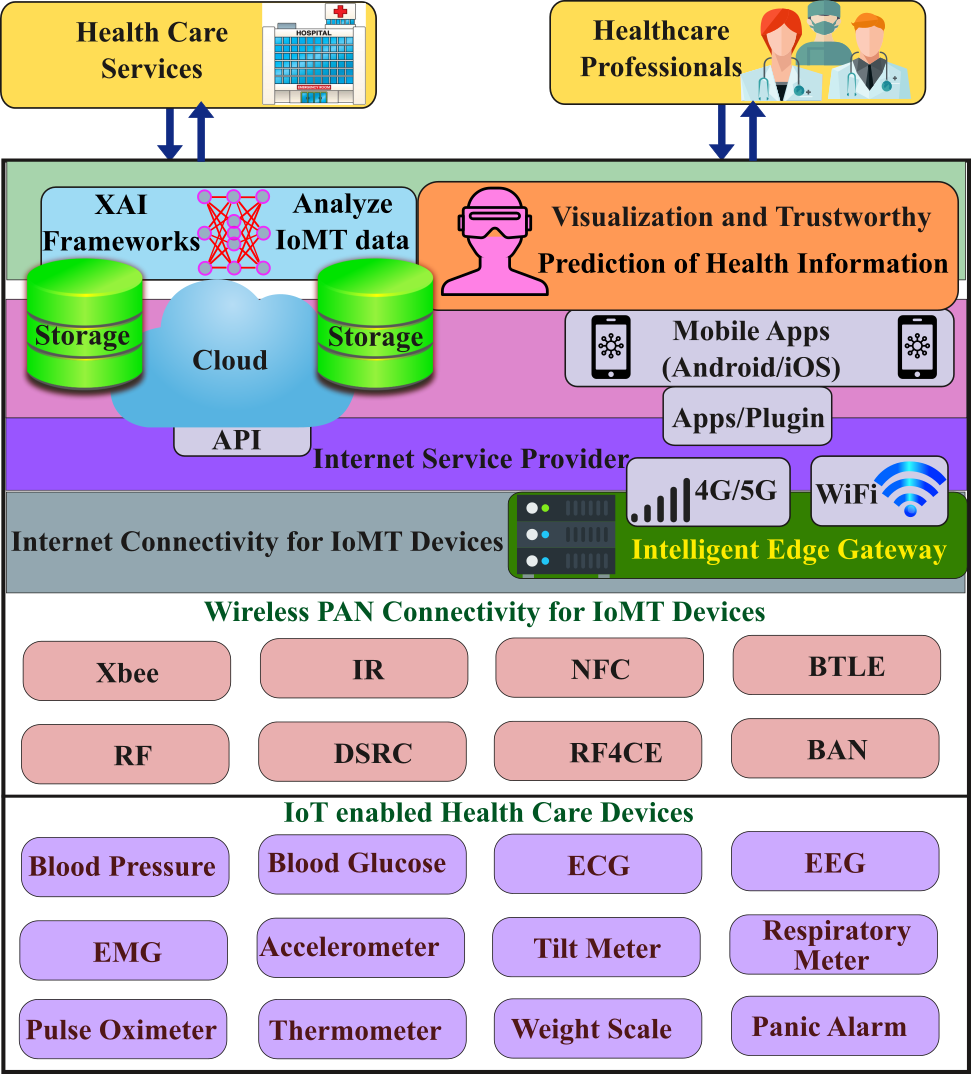}
  \caption{The IoMT Framework using XAI models for trustworthy medical services.}
     \label{fig:IoMTXAI}
\end{figure}

The growth of IoT today has resulted in some exciting advancements in the healthcare sector. Using telemedicine, patients can directly interact with doctors in any part of the world. It has improved the satisfaction of the patients since the time duration they spend interacting with the doctors is higher. In certain cases, there is no need for the patients to visit the emergency room or hospital because of remote health monitoring, which is often known as telehealth. It also helps patients in reducing transportation charges. Home-based telehealth systems can be implemented through IoT without affecting the quality of lifestyle~\cite{ullah2016effective}. Fig.~\ref{fig:IoMTXAI} shows a sample IoMT framework using XAI models for imparting trustworthy healthcare services. To be precise, the connected IoT-enabled healthcare devices are competent for communication using wireless personal area networks (PAN). The decisions made by the XAI frameworks operating on the accumulated data from the IoMT devices provide better assistance to the patients as well as the healthcare professionals. 

In the healthcare information system, the data acquired from the patients is the core requirement, and this volume of data needs to be processed quickly to effectively predict the status of patients~\cite{elhoseny_secure_2018}. IoT devices deployed for medical applications can generate a huge volume of data within a few seconds. The type of data depends on the various healthcare IoT devices accumulated under the framework~\cite{fadhil_hospital_2012}. The provision of a secured latency-free framework for healthcare services helps in reducing bottlenecks and network latencies. However, wireless sensor networks, along with body sensors in healthcare centers, need access points to cover a large area for communication.

The challenges linked with e-healthcare solutions have been well-investigated in~\cite{brito_trends_2016} along with their design and implementation for providing better electronic healthcare schemes. Besides, researchers have put forward cloud computing and networking solutions for e-healthcare with adaptable architectures~\cite{caldeira_intra-mobility_2013}. Biosignals acquired from patients using traditional approaches have been replaced with wearable systems for better monitoring and controlling patients. The use of sensors, such as electrocardiogram (ECG) electrodes, photoplethysmogram (PPG) sensors, along with wireless communication capabilities, provides better-anticipated solutions for ubiquitous healthcare~\cite{chen_data-driven_2018}.

If the currently ongoing research fails to find solutions to address the challenges in healthcare, the problems tend to continue at a fast pace. The solutions to the problems are possible and can be handled by the effective usage of modern technologies. Clinical tests may cost millions of currency. Moreover, the people subjected to such tests have to wait for a long time to gather the required information. Useful healthcare data of millions of people are already available as electronic health records. Healthcare solutions need to be proactive and provide ample solutions before the disease goes to the next stage. Modern healthcare solutions help in preventing many chronic diseases with ease. Moreover, modern healthcare does not require spot-checking. They facilitate remote monitoring of doctors and patients whenever it is necessary. As end-users of the technology, we can gain more out of it through better management and handling of healthcare information.

\subsubsection{AI-driven IoT Systems for Healthcare Support}
AI in healthcare is revolutionizing the medical industry with its great impact. The demand for handling a large volume of data generated for IoMT devices needs proper analysis to make accurate decisions. AI algorithms could handle such a large volume of medical data, which consists of a very vast number of attributes to process and analyze the data. ML and DL solutions are capable of handling high-dimensional medical data. Healthcare industries are getting benefited from AI technology while unwinding a huge set of medical records using cognitive technologies. 

AI-based healthcare mechanisms also suggest that healthcare experts estimate the required medicine, body parameters, and other essentials for providing high-quality health services. AI is also playing a leading role in Health Information Systems (HIS) for handling vital medical and healthcare administrative information of patients in a structured form. Many healthcare systems also use ML and DL approaches for making vital decisions with higher levels of accuracy from the available healthcare information. The AI-based approaches help make better predictions about the patient's health and help the doctors in making decisions. The integration of high-speed communication technologies, such as 5G, into healthcare devices, enables us to provide effective healthcare solutions. 

ML and predictive analytics are widely available across all healthcare domains today. ML models are designed and installed in each product to deliver vital data, so caregivers can now easily make predictive analytics routine and actionable steps forward. It will be valuable access for a caregiver using to uncover patterns of inpatient data through data-driven ML models. The decision tree is one of the popular ML techniques that is meant for grouping data according to a certain criterion, such as age and income of a patient, where he or she lives and his or her diet habits, to estimate the likelihood of these patients to be affected by a certain disease.

In Electroencephalograph (EEG) analysis, a skull cap with a mounted surface electrode is placed on the area of interest, and it records the brain activity (skeletal muscle) in real-time~\cite{liu_recent_2014}, \cite{chen_new_2017}. However, the acquisition of the EEG signal always creates noise, and thus the signal analysis and interpretation are quite difficult. It requires rigorous training to identify the feature which specifies the mental state. Brain-Computer Interface (BCI) is the main driving force for the EEG signal analysis with DL algorithms~\cite{jung_removal_2000},~\cite{schlogl_fully_2007}. Specifically, the interpretation of brain signals is required at an extreme level of precision. DL algorithms are applied to various levels from the segmentation of noise for the classification of signals in healthcare applications~\cite{moretti_computerized_2003}. 

\subsubsection{XAI for IoT-based Healthcare Support Systems}
Even though IoT devices play a predominant part in the healthcare sector, the transparency of utilizing the IoT data for predicting health information and security of IoT data can be enhanced using the XAI concept as an integral part of the healthcare data acquired from IoT devices.

XAI models used in disease diagnosis and prediction could be largely driven by the IoMT devices employed for assessing the health parameters of patients. Further, the fundamental operational procedures involved in the XAI models are very transparent, and they are largely used in IoMT-based disease detection and prediction. Moreover, they are also largely anticipated to provide valid reasoning on the decisions made by the model to the patients. Moreover, the XAI model also helps to address the moral and legal implications involved in the diagnosis of diseases. 

\subsubsection{Lessons Learned}
Even with expert human knowledge, the problem of intra variability persists. Nowadays, computer-based diagnosis is quite popular and accurate. Even the performance measures, such as the classification and identification of accuracy, have been improved by using ML-based diagnosis methods. DL is further improving the accuracy of signal classification with a larger dataset in a minimum time. DL systems can enhance the quality of biomarker assays considering DNA sequences, methylation, gene expression, chromatin profiles, and many other measurements. The vital healthcare-related features from the IoMT data can also be assessed using parameters, such as selectivity and invariance, for both image recognition and gene expression analysis. Further, the usage of Convolutional Neural Networks (CNNs) is the modern visual image processing powerhouse in the field of biomedicine. The outcome from the XAI literature shows the successful application of transparent models in the medical practices associated with the IoMT and smart medical wearable devices. The XAI models thereby enhance the confidence of the users in the AI algorithms of the healthcare sector, as XAI models are applied in a wider range of medical applications. The main lessons learned here are provided as follows.

\begin{itemize}
    \item Data analytics for mobile health and its relevant issues are needed to be carefully managed with appropriate solutions. It can be achieved by the implementation of IoT-enabled systems for real-time health monitoring and performing big data analytics.
    \item  Awareness with respect to existing choices of big data analytics techniques for IoT-based healthcare systems is essential when assessing and selecting a suitable methodology for predicting and making decisions.
    \item The history of patient ailment can be analyzed via data analytics and with connected technology and the information can be accessed at any time and anywhere with cloud computing and IoT. This motivates the development of various handheld smart devices and applications in the healthcare domain. 
    \item IoT devices and networks can be made trustworthy by applying human-centric AI. It enables IoT devices to learn from data and provide interpretable explanations about the carried out estimations or decisions. These explanations can assist humans in detecting non-reliable predictions. Furthermore, the computation costs involved in the XAI model depend only on the training cases, several layers and neurons per hidden layer. Observing the explanation from a larger group of users and IoT devices can ensure auto-generation of explanations in the human-centric AI incorporated into IoT devices. It can be implemented in smart homes and healthcare services, providing reliable and trustworthy explanations of the decisions made by IoT devices. For instance, the depression levels of patients can be tracked, and the XAI model can warn the family, friends, and doctors of the patients with appropriate explanations.
\end{itemize}

\subsubsection{Future Works}
As the sensor manufacturing technologies get more sophisticated and cheaper, it is more convenient to develop smart healthcare devices, such as smart pill bottles, smart wearable devices, devices used for measuring correct postures, and many more. The applications are not just restricted to the development of compact hardware devices. IoMT smart wearable devices can be designed with flexible and lightweight materials to make sure that it is always comfortable to wear. Moreover, the AI frameworks support efficient information sharing between smart medical devices via the aid of appropriate lifecycle management frameworks with improved accuracy. Furthermore, the edge gateway devices close to the network collect those vital medical data through Wi-Fi, Zigbee, Bluetooth, NFC or any other wireless communication protocols. 

People working at home due to the present pandemic situation are often prone to bad maintenance of their health, among which the bad posture of seating is predominant. IoMT devices can be positioned on the neck, and they can assist in tracking the exact position of the head and neck to determine bad posture. It can be used to position the neck and head in real-time using an extremely precise pivot sensor and a three-axis accelerometer. XAI frameworks, in such applications, can precisely observe the health parameters and determine the future impact on such bad postures. The adjustment of frequency and intensity of alerts from the device, guides the users to address their bad posture habits and tracks their progress towards better posture over time. Among all the key challenges, integrity and reliability of the data are particularly significant. Since these are associated with the life and death of the patients, a proper big data governance environment needs to be devised to provide better healthcare solutions \cite{bardhan_predictive_2015}.

\subsection{Industrial IoT}

The latest technology trends in the field of AI and robotics are both important for end-users from different domains. While combining AI and robotics, where the robot is interpreted as the body and the AI as the brain. There is an overlap between these two technologies, in which AI is an area of computer science that helps us to develop computer programs that can learn by themselves such that these data can be fed and they learn from this data, or we human can use sensors and input data to help the algorithms learn by themselves~\cite{malik2021industrial}. Robotics, on the other hand, focus on building and operating robots with mechanical and electronic parts that can do things in an autonomous manner. In the manufacturing sector, for many years, we have had certain tools that can build cars, and now we have very advanced robots that have more sophisticated functionalities. For a long time, we have had robots to build things, such as cars, but they have been dumb robots. These can be programmed to pick something, such as screws, and put them on a wheel of a car, spray paint on the car, but these cannot intelligently make decisions. Nowadays, we can give those robot sensors and cameras that act as their eyes, and we instill intelligence to these robots as of the brain, to construct AI-enabled robots. A drone, for instance a robot, with its brain as AI, can fly in an autonomous manner. We now have self-driving cars again combining robotics and AI tools. 

\subsubsection{The IIoT and Smart Manufacturing}
Beyond the complete integration of sophisticated digital systems in the modern manufacturing sectors, concrete data analytical services are also in larger demand. Further, it is mandatory to establish them to handle the information generated by IIoT systems and smart machines. It leads to actionable insights on the made decision that could deliver a profitable return on investments in the manufacturing process. Thus, sustainable smart manufacturing can be used for smoothly managing the life cycle of the products manufactured associated with the IIoT data, Big Data analytics, and ubiquitous serving facility, as proposed in \cite{ren_comprehensive_2019}. In the smart manufacturing ecosystems, throughput bottlenecks are one of the most common issues that may occur in the production line that generates data from machines in larger volumes. The authors in \cite{subramaniyan_data-driven_2018} implemented an auto-regressive integrated moving average method for predicting the machine behaviors in the active periods with the data acquired through the usage of the bottleneck prediction algorithm. Specifically, the data is channelized and handled effectively to overcome the bottleneck issues and improving the throughput of smooth production tasks. 

IIoT data collected from smart machines enabled with sensors in smart factories are also used for condition-based maintenance \cite{lin_concept_2019}. Specifically, three-stage condition-based maintenance is employed using an ensemble learning algorithm with the first stage involved in training the ensemble, while the second stage detects the imbalanced IIoT data, and the final stage is involved in creating new ensembles. Experimental analysis of the condition-based maintenance provides high accuracy in detecting all the drifts in the data.  

On the other hand, text mining on the IIoT data is another stream of the business prospect that governs Industry 4.0. The text mining data are used for suggesting better operations, technological solutions, improved work skills of the operators, and improved business opportunities~\cite{galati_industry_2019}. Ensuring privacy in data mining is one of the key enablers for protecting massive sets of IIoT data from machines in smart industries. Such large data sets are effectively managed with the minimized transfer of data across nodes using MapReduce frameworks~\cite{keshk_privacy-preserving_2018}. This implies that the IIoT data can enhance manufacturing processes in smart industries along with proper handling and analysis of industrial data.   

\subsubsection{AI-driven IIoT}

Quicker analysis of industrial data and accurate decisions based on the data are made feasible when the 5G communication systems are being paired with AI and Big Data. Technological advancements in the sensor fusion process are proved to produce valuable and accurate information on the processing environment. They can make decision-making tasks easier combining the support of AI and ML. Challenges in data analytics and decision-making tasks are made feasible with the connected industrial devices using the developed e-maintenance framework through sensor fusion techniques~\cite{turner_intelligent_2019}. Moreover, the audit trail for the collection and maintenance of accurate decision-making in smart manufacturing processes are addressed. 

From the 5G networking prescribed for Industry 4.0, network slicing providing the required level of Quality of Service (QoS) needs to be dedicated to ensuring the QoS driven by 5G networks. A new resource allocation scheme is proposed by Messaoud \textit{et al.}~\cite{messaoud_online_2020}, which ensures reliability, delay, and bandwidth allotted for the IIoT devices connected to the network. Subsequently, the developed approach provides a significant reduction, in terms of minimum energy consumption and packet error, for enhancing accuracy in decision support systems, which helps in serving more industrial devices.

Real-time control and reliable sensing, as well as actuation, are realized with the deployment of Tactile Internet in smart industries~\cite{aijaz_tactile_2019}. Moreover, the role of tactile Internet in conjugation with 5G networks for AI and edge-computing is addressed with its ability to provide high-performance, low-latency, and ultra-reliable wireless communication standards for industrial applications. Besides, Tactile Internet is also widely used in the medical industry for observing patient history from remote places~\cite{vora_tilaa_2019}.

\subsubsection{XAI for the IIoT}
\begin{figure}[t]
  \centering    
  \includegraphics[width=0.95\linewidth]{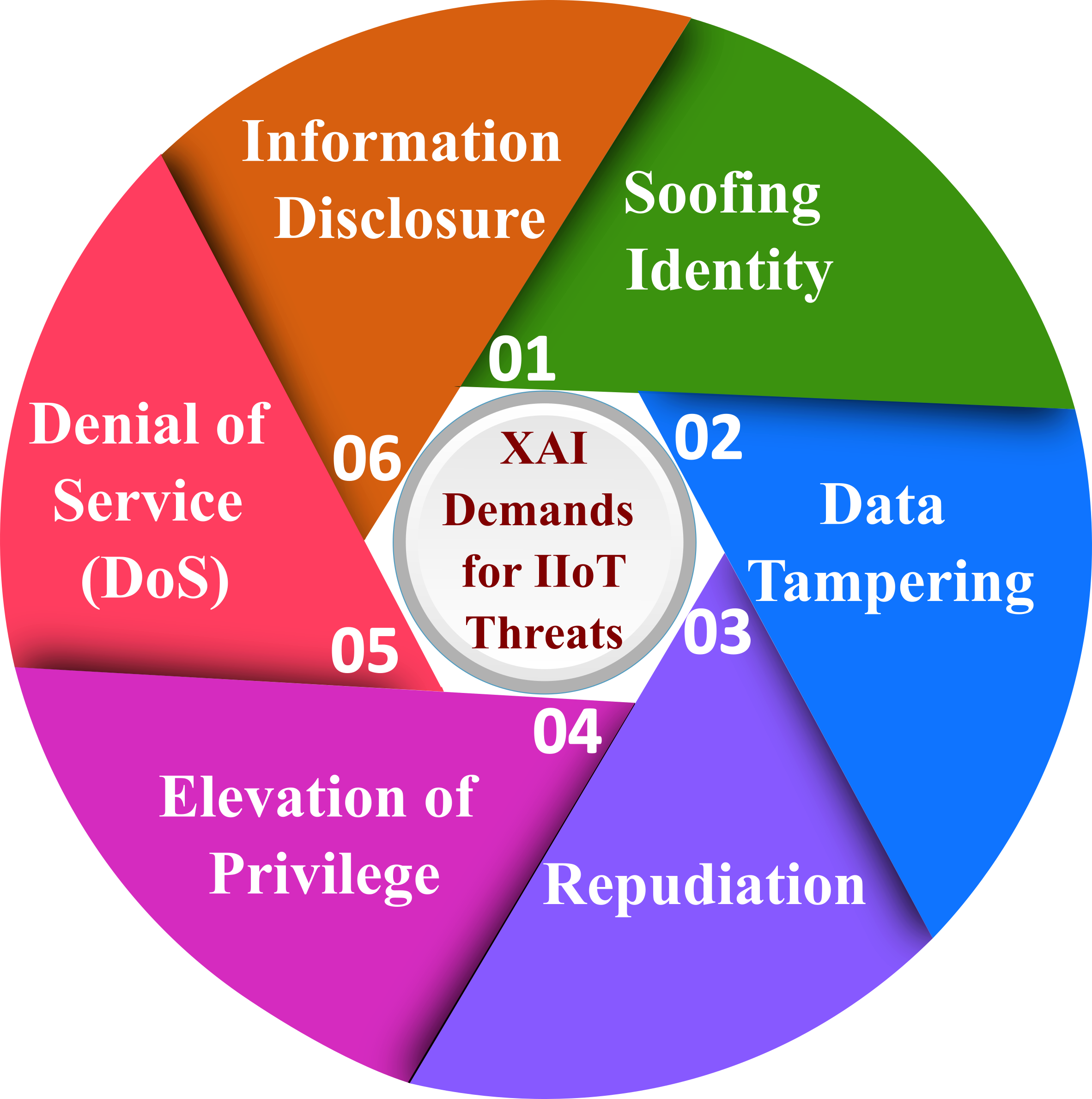}
  \caption{Major IIoT threats that need trustworthiness from IIoT devices / machines.}
     \label{fig:IIoTthreats}
\end{figure}

With the adoption of connected systems in the manufacturing process, we also face increasing cybersecurity risks. Therefore, an adaptation of Industry 4.0 requires close collaboration of Information Technology (IT) experts for the implementation of cybersecurity with the best aspects across the digital ecosystem to enhance privacy and security features. To impart the trustworthiness of IIoT data in smart manufacturing, several efforts have been made. For instance, trust between humans and machines is enforced using cyber-physical-human analysis \cite{jiao_towards_2020}, which enhances the trust in dynamic modeling, cognitive prediction, and optimized interaction between humans and machines. Some of the major security threats faced by IIoT devices/machines are shown in Fig.~\ref{fig:IIoTthreats}. Similarly, Hehenberger \textit{et al.}\cite{hehenberger_design_2016} reviewed the mechatronics systems in the industries and their transition towards secured CPS and cloud-based IIoT systems. Secure Blockchain Tokenizer is also used for securing IIoT data collected from the supply chain in smart industries\cite{mazzei_blockchain_2020}. Data acquisition, monitoring, and teleoperation of IIoT data from CNC machines are performed using an Internet-based client-server model through the web systems \cite{alvares_development_2018}. The developed CyberDNC web systems are validated with a focus on secured connectivity between the servers and the CNC machines.  
Moreover, MCloud platform was developed by Lu \textit{et al.}~\cite{lu_energy-efficient_2019} for establishing a secure and energy-efficient production network with improved privacy features and enhanced flexibility in the Industry 4.0 environment. The framework was also used for collaborative energy-efficient manufacturing for distributed manufacturers to coordinate production processes that are customized based on requirements. In \cite{liu_industrial_2020}, the researchers developed an industrial blockchain-based product lifecycle management model for the exchange of secure data services, secured storage platform, decentralized data access, and interoperability among Industry 4.0 compatible networks. IIoT based data-driven security enforcement is also applied to the production of circuit breaker~\cite{chen_framework_2020} in smart factories. 

\subsubsection{Lessons Learned}
Clearly, as the digitization and social trends in modern industries are getting evolved, the expectations of consumer segments are far beyond the current digital transformation. We explored the requirements to monitor, operate and control appliances in industries with fewer efforts using XAI models. Some key outcomes learned from this section are provided as follows.

\begin{itemize}

    \item The evolution of IIoT has made a paradigm shift to meet expectations through smart industries and appliances. Smart devices can communicate effectively in the industrial environment with modern wireless communication technologies employed in IIoT-enabled machines \cite{khan2020industrial}. IIoT, being a subset of IoT, is supported with numerous enabling technologies for the disruptive evolution of digitization in smart industries\cite{younan_challenges_2020}. Sustainable technologies for Industry 4.0\cite{ghobakhloo_industry_2020}, socio-technical view of modern industries\cite{beier_industry_2020} and state-of-the-art tools for smart machines\cite{xu_industry_2018} are summarized by many researchers, and the significance of IIoT towards social trends is compiled in the manufacturing sector. 
    
    \item Few researchers summarized the case studies of using IIoT in modern industries\cite{lu_oil_2019} with the potential requirement, involved challenges \cite{silva_implementation_2020} and new advancements of Industry 4.0\cite{hervas-oliver_place-based_2019} driving the paradigm shift of industrial manufacturing processes. Digitization of industries also invokes the demand for the integration of CPS with IIoT\cite{kipper_scopus_2020} and improved cognitive ability in solving challenges at socio-economic smart industrial environments\cite{zolotova_smart_2020}.  

    \item Adopting Industry 4.0 requires both vertical and horizontal data integration across various business segments. Vertical digitization includes procurement, product design, manufacturing, supply chain,  product life cycle management, logistics, operations and QoS, all integrated for seamless flow of data~\cite{faheem2018mqrp}. Horizontal digitization includes data integration with suppliers, customers, and key partners. Achieving integration requires upgrading or replacement of equipment, networks, and processes until a seamless digital ecosystem is established.  

\end{itemize}

\subsubsection{Future Works}
Global IoT technology is expected to rise to 6.5 trillion USD by the end of 2024 with the disruption in wireless communication services. To be in the race of this disruption in the industrial domain, wireless communication services demand a centralized computing approach and thereby provide support towards self-healing, automated and optimized infrastructures. Cloud/Centralized Radio Access Network (C-RAN) and SDN are also in the race to provide novel technological trends for 5G networks. They provide faithful coordination of Inter-cell and multi-point interference management with high throughput, ultra-reliable, and low latency features\cite{zanferrari_morais_when_2020}. New development in the field of IoT in synchronization with big data, cloud computing, and AI have triggered the disruption in wireless services for industrial applications \cite{zhang_2020_2020}. Cyber-physical production is gaining popularity in smart factories with the support of computer-aided tools, robots, and other technological advancements, providing flexible manufacturing processes. This enhances the supply chain, distributed warehousing, and automation process through the support of 5G communication services in smart industries\cite{vernadat_information_2018}. Industry experts think that Industry 5.0 is a new revolution and 6G will be a new wireless generation~\cite{maddikunta2021industry}, but researchers believe that this transition will not halt at 6G. Rather, it will continue beyond 5G and 6G as many technical challenges are yet to be solved.

\subsection{Internet of City Things}

\begin{figure}[t]
  \centering    
  \includegraphics[width=\linewidth]{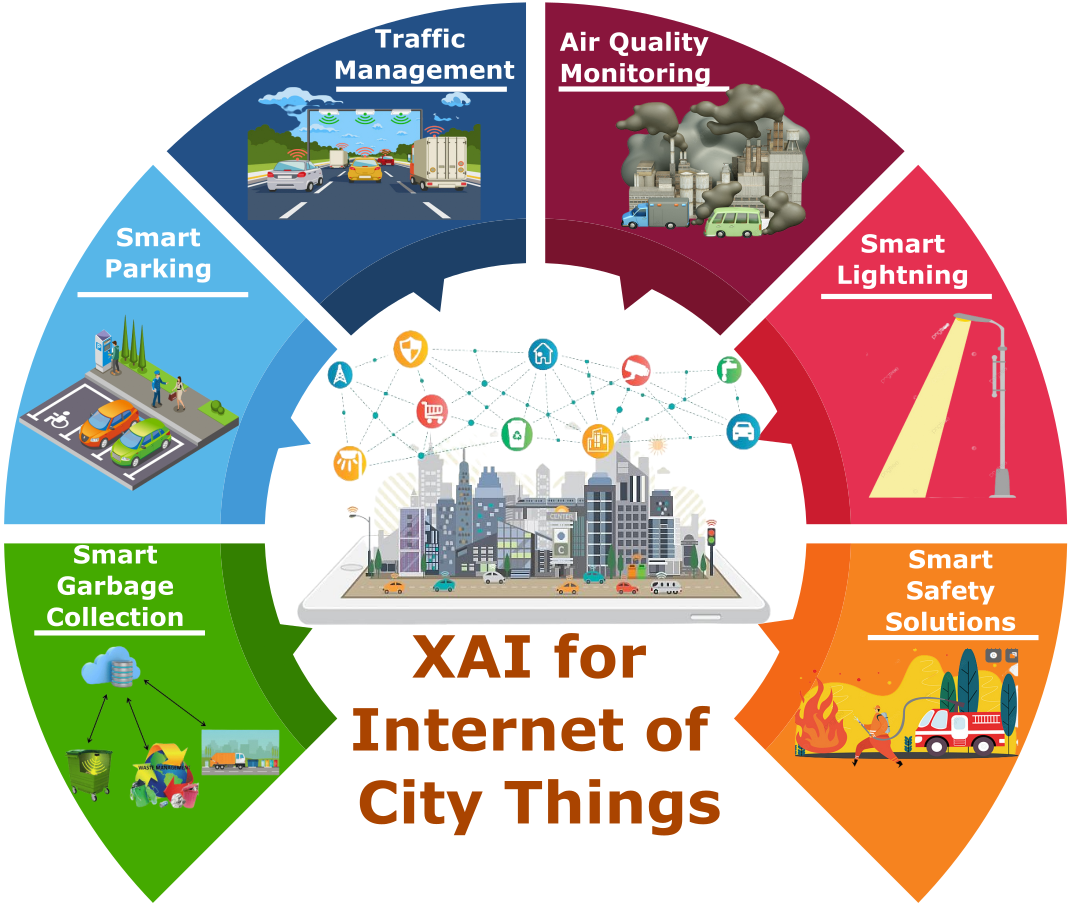}
  \caption{Crucial IoCT sectors seeking XAI attention.}
     \label{fig:IoCT}
\end{figure}

Over the past 50 years, the percentage of people who live in cities across the world has doubled and almost two-thirds of the global population will live in cities in the forthcoming decades. Urban and rural areas are evolving to cater to the changes in technology while presenting new opportunities for improved public safety and connectivity and for the overall experience of the visitors and residents living in smart homes. In a smart home or smart building, the devices are connected via the Internet for provisioning remote monitoring and management of home appliances. Smart home technology, also known as the Internet of Home Things (IoHT), provides the house inmates convenience, security, scheduled maintenance, and energy-efficient control. It allows them to control smart home appliances with ease from remote places using apps on their smartphones or other networked devices (e.g., grant or deny home access through smart locks and check-in security cameras). Fig.~\ref{fig:IoCT} shows the major focus of XAI for IoCT applications.

\subsubsection{Smart City IoT Products}
In the present landscape of the IoCT, the IoHT has become its integral component. Smart cities are envisaged from the Infrastructure perspective to have four pillars, including Physical, Social Infrastructure, Economic, and Institutional Infrastructure. The residents in the smart cities are the focal attention of each of these pillars. A smart infrastructure framework was developed in~\cite{park2018design} for a disaster management system using IoT and Augmented Reality based smart buildings. To be precise, the authors introduced a large category of IoT devices that can be deployed for smart industries along with an interoperable test-bed used in the smart cities for testing the Augmented Reality-based disaster management service. Furthermore, this system also provides a safe and quick means of rescue guidelines for the residents in the smart buildings during emergency situations.  

In addition, smart city services, such as the management of energy, water, and waste materials, are also driven by IoCT devices. They help to address the indispensable challenges imposed by rapid urbanization. Further, for optimized usage of resources and for robust decisions from IoT devices, their integration with AI services are gaining popularity among the end-users.

\subsubsection{AI driven Smart City IoT Products}
AI-powered solutions also contribute greatly towards the deployment of smart appliances for homes, buildings, and other city infrastructures. In smart buildings, multiple vital parameters, such as lighting, security, and energy thermostats, need to be integrated into a secured and cost-effective solution. The technological convenience and efficiency of AI algorithms can play a vital role in the integration of smart cities and ensure appropriate decisions on the security issues, energy consumption, and demands of the appliances. In \cite{englund2021ai}, traffic control operations in smart cities are automated using AI frameworks for classifications of vehicle functionalities in traffic. Further, it assists in enhancing the smart cities and the operations of communities in the data-centered society. However, as the number of interconnecting IoCT devices increases in smart cities, the AI frameworks must possess additional features to explain the decisions made by the AI models, which are rare in conventional AI models for smart cities.

\subsubsection{XAI for Smart City IoT Products}
Many governments across the globe are racing to integrate AI and other modern technologies into every operational aspect of cities under their control. Most of the parts include public transportation, IT connectivity, power, water supply, sanitation, waste management, and e-governance. The technological implications of Big data, IoT, and AI make the deployment of smart cities feasible in these aspects. 

Interference alignment-based communication security in IoT devices for smart cities imparts a secure green communication framework. In this scheme, the artificial noises and information signals are meant to harvest energy for establishing green communication \cite{xie2020secured}. In this work, the transmission rate is not affected by the artificial noise and improves the secrecy performance with good efficiency in energy harvesting tasks. XAI frameworks on green communication help to estimate the possible measures considered for aligning and mitigating the interference in the IoT communication and assist in exploring the parameters being optimized for energy harvesting and management.

Big data analytics plays a vital role in smart city applications. Further, those applications can process data from IoT devices and sensors to recognize patterns and demands of the people and infrastructures present in the environment. Employing robust and secure data analytics can considerably stop the congestion on roads, reduce accidents and congestion, and help drivers find many places to park their vehicles. Big data can also improve smart lighting, improve energy as well as utilization of water and reduce crime. In \cite{rathore2018exploiting}, the authors exploited IoT and big data analytics of smart cities using real-time urban data and subsequently used them to digitize the conventional technologies present in cities. Further, this work \cite{rathore2018exploiting} implements city Planning using big data analytics and further processes traffic information to alert the drivers. In addition, the secure and scalable real-time processing of data from smart cities is utilized for enhanced user experience to make human lives easy with the assistance of IoT devices interacting with the environment.

Deployment of CPSs in smart city applications is poised to significantly improve the living of people in the cities. However, these are prone to vulnerability and risks of attacks on the IoT devices employed for monitoring and automating the tasks in the cities. Eliminating security vulnerabilities in smart cities involves the role of both technology and government entities. In \cite{habibzadeh2019survey}, the authors explored the possibility of managing the cybersecurity issues, data privacy challenges, and policy issues in CPS deployments of smart cities. Further, the theoretical and practical aspects of cybersecurity and privacy can be well addressed using AI frameworks through ML and DL techniques. Popular works on connected cars \cite{olufowobi2019connected} and E-governance \cite{yang2019privacy} have employed ML and DL techniques for ensuring secure frameworks for the IoT systems in smart cities. Furthermore, the XAI models on the frameworks ensure the interpretation of the decisions to provide robust and secured services. 

\subsubsection{Lessons Learned}
From the previous discussions and a few other prospective end-use cases of imparting trustworthy solutions for smart cities, we have summarized that XAI can be an effective tool in providing confidence, secure and trustworthy usage of the data from IoT devices that need to maintain a fair means of communication and exchanging data among these devices in the network. The key benefits of the mentioned applications are brought forward by the XAI models that provide human-centric AI in IoT devices, which further enable the XAI system model to accurately predict and provide trustworthy explanations of the decisions made by the models. This, in turn, leads to a better coexistence between the IoT devices and XAI models owing to the predictive ability of the models deployed in smart cities. They are largely associated with the ability to use the learned data to determine crucial prospects and thus provide better resource management solutions. The key lessons learned here are given as follows.

\begin{itemize}

    \item High-speed communication and trustworthy data transfer are key requirements in smart city applications. In the aforementioned applications, the IoT devices in the network demand reliable technology for high-speed connectivity and thereby trigger quick prediction by the XAI models. Further, it is needed to determine the trustworthy solutions of the made decisions, which can be thought of as a series of the optimized outcome made to select the best and quick path in the data transmission. In this context, from the perspective of smart city construction, reliable communication services are in demand, and hence better solutions for building reliable and robust backbone communication networks are required. 

    \item Even though providing the optimal energy-efficient smart solutions in the aforementioned applications is quite challenging, and these provide guaranteed sustainable ways of transforming the IoT devices for reliable and long periods of fault-free operations in the automation processes. The potential impact of energy management in intelligent smart buildings ensures the foundational aspects of the IoT devices before additional solutions driven by robust XAI models and services can arise. XAI models in these perspectives can determine the sectors where the energy is spent and provide awareness of the energy-saving phenomenon and its optimized usage in smart city applications.
    
\end{itemize}

\subsubsection{Future Works}
The promising rise of IoT usage in smart cities integrated with AI technology was expected to bring data-centric solutions for addressing urban challenges. Whilst a few continents were ahead of the curve, most of the legacy cities are trying to upgrade their age-old infrastructure to build smart cities. To make the cities smarter, more efficient, and more sustainable for their residents, the XAI models can provide adorable solutions to the decisions made by the smart IoT-based devices employed for the growth of smart cities. Data analysts argue that there are four infrastructure investment opportunities (such as Buildings \& construction, Energy and Water \& waste management, and Enabling technologies) for the sustainable development of infrastructures with the support of XAI solutions for smart cities. The accelerated development of using XAI models in these sectors will lead to massive trust in the deployment of feasible solution-driven IoT devices for the growth of smart cities. Although the technological impact of XAI, big data, and IoT have received considerable attention, there are still many practical issues related to the implementation aspects of XAI models in IoT systems for smart cities, such as the hardware design, security, control in deployed locations and resource management, which require attention from the research community.

\subsection{Summary}

In summary, with the inherent capabilities and maturity of the XAI, it paves the way for transforming the application domains of IoT to impart trustworthy services in association with other enabling technologies. Such technologies promise to offer a flexible deployment of IoT infrastructure and operational improvements as well as facilitate a transparent view of the decisions made by the smart devices to the end-users. However, as the XAI and IoT technologies are getting mature, several new research challenges are emerging that the research community needs to address. This subsection provides a summary of the advantages, challenges, and limitations of interoperability among XAI-based IoT applications. 

\subsubsection{Advantages of XAI based IoT Systems}
The ever-increasing popularity of blending AI with IoT has already been adopted by many business giants. However, the competitive advantage in real-time decision-making can be feasible for them if they tend to explore the mechanism of making decisions by the AI systems integrated with IoT frameworks. Even as the core functionalities of IoT are to make use of the stream of data accumulated from the sensors mounted onto end users, such as smart machines and other environmental monitoring systems through internet connectivity, its inevitable final stage is the data analysis phase for extracting patterns from the data. Despite the crucial role played by AI models for extracting these patterns from the data generated by IoT devices, these lack in exhibiting the stages of decisions to the end-users. Deployment of XAI models can unlock these features of exploring the decision-making phases of the systems based on the data generated from the IoT devices.

\subsubsection{Interoperability challenges and limitations of XAI based IoT Systems}
The maturity of any technology comes with the cost of solving the involved challenges. Specifically, about interoperability in the IoT interface, few challenges exist due to the coexistence of multifarious systems, devices, sensors, and equipment that interchange location, time-dependent information in various data formats, languages, data models, construction, data quality, and complex interrelationships. Multi-version system designers and manufacturers, over time, under varying application domains, formulate the conditions of global agreements, and hence commonly accepted specifications are quite difficult.

\section{XAI: Cutting Edge Developments for Dependable IoT Systems}
\label{sec:XAIforIoT}

The services provided by the XAI models for dependable IoT systems aim to drive through certain value add-ions. Although upgrading the features of XAI is independent of the timeline of using 6G services and edge XAI structures, these are crucial for XAI Security Enhancement and hinder the developments towards integrating XAI with IoT. This section envisages and summarizes the prospective cutting-edge developments in XAI for IoT systems. A few of the most popular cutting-edge developments are summarized in Table~\ref{tab:cuttingedge}.

\begin{table*}[!ht]
\centering
\caption{Existing Works on Cutting Edge Developments for Dependable IoT Systems} 
\label{tab:cuttingedge}
\begin{tabular}{|p{0.75cm}|p{0.5cm}|p{6.0cm}|p{9.0cm}|}
\hline
\textbf{Ref.} & \textbf{Year} & \textbf{Key Inference} & \textbf{Cutting Edge Developments}\\ \hline
\hline
\cite{hussain2020machine}  & 2020 & Resource management in IoT networks. &  
\begin{itemize}
    \item ML and DL at the edge devices.
    \item Reduces extra bandwidth.
\end{itemize}
\\ \hline
\cite{yao2019recommendations}  & 2019 & Management of decision-making critical issues. &  
\begin{itemize}
    \item Engaged in on-device training.
    \item Manages the sensor data from all the IoT devices.
\end{itemize}
\\ \hline
\cite{wang2022distributed}  & 2022 & Deals with a large volume of data. &  
\begin{itemize}
    \item Distributed RL provides a viable and effective solution.
    \item Optimizes training time and maximizes the throughput.
\end{itemize}
\\ \hline
\cite{temesgene2020distributed}  & 2020 & Focuses on energy harvesting in the network &  
\begin{itemize}
    \item Virtualizes cells and controls their power consumption. 
    \item Distributed Deep RL is incorporated for harvesting energy.
\end{itemize}
\\ \hline
\cite{grossberg2021attention}  & 2021 & Evaluation of cognitive states of human brain.  &  
\begin{itemize}
    \item Distributed supervised learning models are used for feature extraction.
    \item Attention features study the cognitive states of the brain.
\end{itemize}
\\ \hline
\cite{rapp2020distributed}  & 2020 & Heterogeneous distributed resources gains benefit from the distributed learning frameworks. &  
\begin{itemize}
    \item Provides an optimal topology that fits the capabilities of the system.
     \item Significant rewards from the strongly contributing IoT devices. 
\end{itemize}
\\ \hline
\cite{tan2021towards}  & 2021 & Personalized FL. &  
\begin{itemize}
    \item Guidelines towards the development of personalized FL.
    \item Realistic PFL approaches are highlighted.
\end{itemize}
\\ \hline
\cite{hsu2020privacy}  & 2020 & Malware detection in edge computing applications.  &  
\begin{itemize}
    \item Privacy-preserving FL classifier was developed and trained.
    \item Guarantees better privacy features for Android applications.
\end{itemize}
\\ \hline
\cite{ur2020towards}  & 2020 & Reputation awareness for ensuring trustworthy and collaborative training of IoT data. &  
\begin{itemize}
    \item Fine-grained FL is used for decentralizing the ML models.
    \item Blockchain-based scheme used for training IoT data at the mobile edge devices. 
\end{itemize}
\\ \hline
\cite{marulli2020security}  & 2020 & Fairness of the FL models in edge and cloud environments. &  
\begin{itemize}
    \item Blockchain-based security-oriented architecture is employed 
    \item Context-aware synchronization of models for IoT is ensured.
\end{itemize}
\\ \hline
\cite{guo2021enabling}  & 2021 & IoT-enabled applications with technical requirements in incorporating 6G services. &  
\begin{itemize}
    \item Realization of 6G energy-efficient components.
    \item Lead role of IoT devices as personal digital assistant.
\end{itemize}
\\ \hline
\cite{chen2020towards}  & 2020 & Convergence of 6G and IoT services for Radio-over-Fiber systems. &  
\begin{itemize}
    \item XAI-based solutions for optic fiber communication.
    \item Provides better spectral efficiency.
\end{itemize}
\\ \hline
\cite{gupta20216g}  & 2021 & Impact of 6G-enabled intelligence in the edge nodes. &  
\begin{itemize}
    \item Ultra-reliable low latency applications for IoT and autonomous vehicles.
    \item Improves the efficiency of the communication services.
\end{itemize}
\\ \hline
\cite{wu2020cloud}  & 2020 & Integrates cloud and edge computing platforms for IoT applications. &  
\begin{itemize}
    \item Scalable solutions for distributing valuable resources.
    \item Reduce the communication delay and storage requirements.
\end{itemize}
\\ \hline
\cite{yuan2021noma}  & 2021 & Impact of 6G-based massive IoT for multiple access. &  
\begin{itemize}
    \item Key performance indicators and their dimensions are summarized.
    \item Joint optimization of channel coding and linear spreading. 
\end{itemize}
\\ \hline
\cite{niu2020green}  & 2020 & Green communication services for IoT.  &  
\begin{itemize}
    \item Better services, mobility, and resource management.
    \item Ubiquitous intelligence in the current communication standards.
\end{itemize}
\\ \hline
\end{tabular}
\end{table*}


\subsection{Edge XAI Structures}
When more IoT devices are connected to the cloud services through a network, scalable operations in the deployed environment could be challenging. Further, it might start to run into physical limitations in network bandwidth as the network gets crowded with data from all the IoT devices. It leads to paying more than the intended cost to the network service providers for that extra bandwidth. To address the resource management in IoT networks, implementation of ML and DL at the edge devices are motivated in \cite{hussain2020machine}. For mission-critical multi-domain operations, the importance of fair, unbiased, trustworthy, and explainable outcome are crucial, which could be achieved by reliable and trustworthy AI algorithms \cite{rawat2021secure}. Among various requirements and demands of IoT devices, proactive discovery is essential in decision-making tasks. This critical issue could be managed by acquiring knowledge from the diversified class of IoT devices engaged in on-device training and equipped with decision-making capabilities \cite{yao2019recommendations}. Owing to the idea of edge computing, our local machines or servers assist in managing the sensor data from all the IoT devices in an environment. 

Furthermore, edge XAI provides techniques to understand better and validate the mechanism of AI models and their decision-making process. Thus, the generic AI structures cannot be utilized entirely solely with the models for imparting interpretable features. Accordingly, certain learning approaches are discussed to frame appropriate XAI structures that could cover the gaps in conventional structures.

Wells et al.~\cite{wells2021explainable} highlighted the limitations of XAI in the Reinforcement Learning (RL) frameworks by highlighting the visualizations, policy challenges, and query-driven explanations. It ensures the enhancement of the decision-making ability of the RL techniques in various ranges of IoT applications. Further, the challenges in provisioning understandable explanations and their prevalence with respect to IoT applications were explored. In~\cite{chiaroni2020self}, the authors investigated self-supervised learning techniques for the perception of autonomous vehicles. Here, the focus was on the analytical methods and the handcrafted designs, which help to represent appropriate scenes in the perceived environment. The learning techniques adopted in this setting could assist mission-critical applications with the aid of XAI frameworks deployed to extract meaningful insights from the perceived data.

\subsubsection{Distributed Reinforcement Learning} 
RL is a common approach for learning from the environment based on the agents taking action and collecting cumulative rewards. However, this learning approach can be expensive and consumes more training time due to its resource constraints, especially when dealing with a large volume of data~\cite{luo2022reinforcement}. Distributed RL is a viable and effective solution, which optimizes the training time, and maximizes the throughput and concurrency process for large datasets. Additionally, this learning technique maximizes the batch size in data transfer~\cite{wang2022distributed}. 

Temesgene et al. \cite{temesgene2020distributed} considered energy harvesting in the network of virtualized small cells and controlled their power consumption using Distributed Deep RL (DDRL) framework. The work in \cite{toromanoff2020end} demonstrated the effectiveness of end-to-end model-free RL for urban driving by utilizing implicit affordability. It was analyzed by considering the pedestrians, traffic lights, and vehicle avoidance issues. Moreover, in \cite{hayes2021practical}, the real-world decision from a multi-objective perspective was studied using RL and planning strategies. In another similar work presented in \cite{dulac2021challenges}, for continuous control of real-world environments, RL and other state-of-the-art learning algorithms are compared, and subsequently, an open-source benchmark standard was put forward. In particular, the work in \cite{feriani2021single} studied multi-agent distributed RL algorithm for AI-enabled wireless standards in coexistence with the model-based RL and cooperative model-based RL while utilizing them for Unmanned Aerial Vehicle (UAV) networks \cite{yang2019energy} and mobile edge computing applications \cite{yang2019energyuav}. Meanwhile, the Distributed ML focused on wireless networks helps analyze the constraints related to computation cost, rate of convergence, scalability, and optimization issues \cite{hu2021distributed}. Further, they could also assist in addressing the adversarial attacks in their end applications. To reduce the complexity of the learning model, the authors in \cite{chen2021communication} proposed a policy optimization strategy for distributed RL algorithm, which then can be used to reduce the communication overheads and establish distributed learning facilities by fetching the best rewards. 

It is worth noting that to use distributed RL effectively, approaches in XAI edge computing, the power, resources, and computation cost need to be optimized concerning the design metrics of IoT devices while meeting the relevant constraints. Although the joint optimization of various parameters is a challenging task, it can significantly enhance the performance of edge XAI structures developed for IoT applications. 

\subsubsection{Distributed Supervised Learning} 
Supervised learning approaches help to separate data through classification mechanisms and fit the data to a particular class through a regression mechanism. While handling a large volume of IoT data from diversified categories of smart devices, improved performance and high-speed decision-making process could be ensured if more such learning models are deployed in a distributed fashion. Distributed supervised learning provides a practical implementation of large-scale distributed services for efficient parallelization in processing and training data \cite{verbraeken2020survey}. It ensures effective interpretation of a bunch of distributed models rather than exploring the decision-making capability of a large centralized supervised learning model.

Grossberg et al. \cite{grossberg2020toward} presented a review on the state-of-the-art of autonomous adaptive algorithms from the perspective of AI and human brain functional capabilities (e.g., cognition, perception, and action) in autonomous intelligent systems. Further, in this work, the distributed learning capabilities of the human brain are contrasted with the adaptive intelligent systems. In another recent work \cite{grossberg2021attention}, cognitive states of the brain are evaluated, which could be effectively imparted as cognitive attention features in distributed supervised learning models. To enhance the performance of distributed ML networks, the authors in \cite{malandrino2021network} proposed a supervised learning approach considering the learning and information nodes for learning cooperation, thus reducing the learning cost. 

Considering the critical aspects of distributed supervised learning approaches, XAI compatible edge developers can deploy intelligence, such as cognitive features, and cooperate with the learning tasks at the XAI edge through the significant sharing of resources in the distributed learning architecture.  

\subsubsection{Federated Learning} 
Federated learning (FL) helps to train AI algorithms across multiple devices, including smartphones, IoT devices, and other distributed cloud services. It allows users to build a model that drives towards data privacy concerns and handles bandwidth issues of IoT devices and smartphones through local training and then sharing the model with the servers \cite{pham2021uav}. The server then combines the models acquired from multiple devices into a single federated model, and this particular model does not have direct access to the training data. 

Practical implementation of FL in IoT devices helps ensure data privacy and reduces communication costs. Moreover, FL possesses its inherent privacy preservation nature, which is widely used in 6G services \cite{liu2020federated,xiao2020towards}. Comprehensive information on different aspects of FL systems, their categorization, and their usage mechanism for data privacy and protection are provided in \cite{li2019survey}. Recent work is \cite{tan2021towards} in this regard, where guidelines are provided for the development of Personalized FL (PFL). Specifically, the PFL techniques are categorized as model-based and data-based methods, and the key ideas and realistic and trustworthy PFL approaches are highlighted. Zhang et al. \cite{zhang2021federated} used a federated transfer learning method for fault diagnosis in industrial machines that ensures data privacy using deep adversarial networks. 

FL has notably secure means of preserving the data from IoT devices and supports local training of data. However, one should note that the privacy-preserving accuracy of the IoT data is determined via the combination of established communication, edge infrastructure, and other device-specific parameters.    

\subsubsection{Blockchain Based Distributed Learning} 
Blockchain-based distributed learning provides a secure framework to eliminate trusted third parties involved in the process of content transactions \cite{nguyen2021federated}. Instead of using centralized service from a single party,  a group of peers and organizations, using the distributed learning technology, are involved in processing the submitted transaction requests digitally signed by the IoT devices involved in the transaction process. 

In a recent survey \cite{otoum2021preventing}, the blockchain-assisted AI-enabled networks, prevention, and control of epidemics are analyzed in a distributed and cooperative healthcare framework. This involves a secure exchange of medical information at the edge device that lies among different healthcare networks. In addition to other viable means of adding edge intelligence, distributed Multi-Armed Bandits (MAB) model \cite{chen2021distributed} are used for Multi-agent decision-making in edge structures via a robust lightweight learning algorithm that allows changes in the topology of a dynamic IoT network. In a network of IoT healthcare devices, blockchain smart contracts are employed to manage the edge training plan using the FL approach, which ensures authentication and trust management of IoT devices \cite{rahman2020secure}. Furthermore, this work employs a lightweight hybrid FL framework and supports full encryption on the collected data with the edge nodes having additive encryption ability on the IoT data, thus providing secure means of IoT health care services. 

It is remarkable that all the possible solutions for deploying XAI structures at the edge of IoT networks so far show that the learning frameworks should be secure, trustworthy, and capable of executing AI algorithms in resource-constrained devices. This will require innovative measures for the efficient implementation of trustworthy XAI structures. 

\subsection{Potential to Meet 6G Requirements for IoT}
\begin{figure}[t]
  \centering    
  \includegraphics[width=0.85\linewidth]{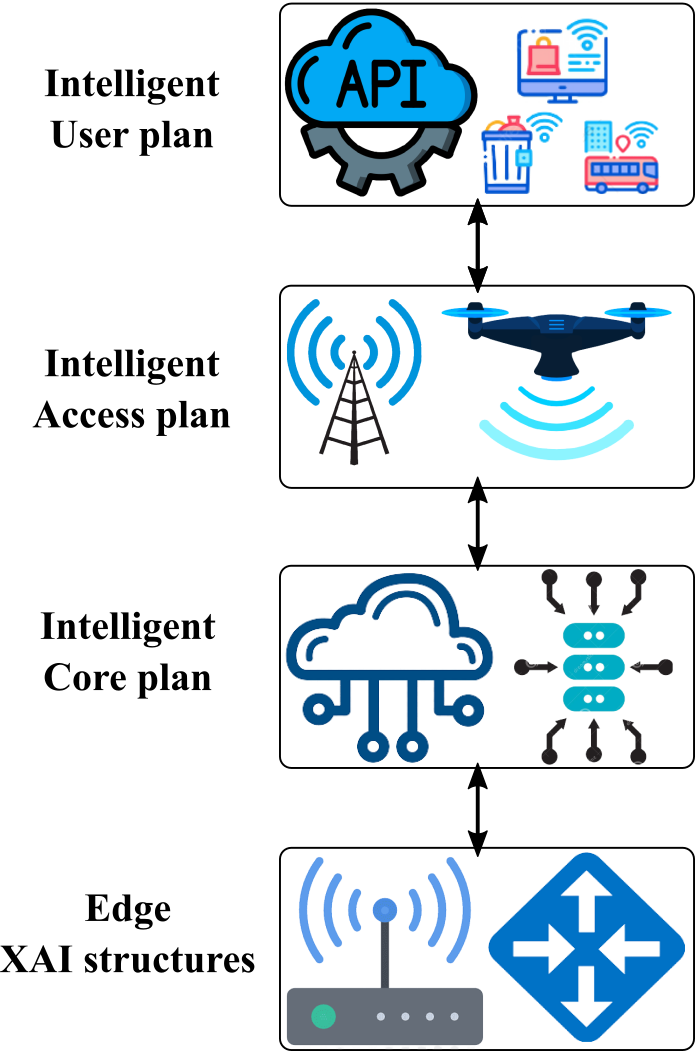}
  \caption{6G architecture for XAI driven IoT applications.}
     \label{fig:6GXAI}
\end{figure}

To date, a plethora of research in wireless communication has provided 6G services to evolve terrestrial, air, and maritime communication into new dimensions. The key players of this 6G technology work under sub terahertz (THz) and THz frequency communication media. There is much bandwidth available at these higher frequencies; therefore, data rates up to one terabit per second seem feasible \cite{zhao2020comprehensive}. As this is one of the key aspects of the 6G communication standard, this can be a better component for infrastructure IoT, ubiquitous sensing, distributed identity, and embedded sensing applications. Guo et al. \cite{guo2021enabling} presented a survey on 6G-enabled massive IoT that summarizes the IoT-enabled applications along with the core technical requirements for incorporating 6G services with new network architecture. Furthermore, the latency in 5G services in milliseconds could be much reduced to microseconds by enabling much denser IoT connectivity. Moreover, the real digital ecosystem could be realized in 6G energy-efficient components. It could also be a hybrid ubiquitous network, which could incorporate all advanced contemporary technologies. Further, 6G for mobile handsets and IoT devices will be playing a lead role as personal digital assistant for individuals. Optical communication services and security features are expected to be much superior in 6G services. Fig.~\ref{fig:6GXAI} shows the layer-wise architecture of 6G services for delivering trustworthy XAI-driven IoT services.

\subsubsection{Massive ultra-reliable, low latency communications (mURLLC)} 
With the requirement of establishing robust end-to-end security and achieving a target latency of around one microsecond, 6G services can be a perfect reliable standard. Establishing reliable communication among an immense number of mission-critical IoT devices forces a drastic change in existing communication standards to ensure channel availability and avoid congestion in the network. 6G communication services providing Massive ultra-reliable, low latency communications (mURLLC) will be a perfect candidate for latency-sensitive IoT networks. As IoT devices demand low-power transmission as one of their key priority, 6G communication standards can address the trade-offs among throughput, reliability, and latency that are prevalent issues in IoT communication. In a recent work reported in \cite{gupta20216g}, the authors summarized the impact of 6G-enabled intelligence in the edge nodes of the network for ultra-reliable low latency applications targeted for IoT and autonomous vehicles. This 6G-enabled edge intelligence improves the efficiency of the communication services in terms of network mobility and communication latency \cite{guo2021enabling,gupta20216g}. 

\subsubsection{Scalable architecture} 
Most of the existing work focuses on customizing wireless services for IoT devices and individual network entities based on static edge computing in a single edge server. However, in the future 6G communication services, a larger number of IoT devices need to be integrated. This requires a decomposable and scalable architecture that could support parallel computing among multiple edge devices~\cite{guo2018scalable}. Such edge intelligence provided in a distributed manner can support better computation and communication services. For future 6G-enabled IoT applications, the role of scalable architecture is expected to play a crucial role. 

Scalable architecture coupled with FL-based 6G services significantly reduces the security threats and enables distributed availability of computation resources \cite{yang2021federated}. In \cite{xiao2020towards}, the authors summarized the convergence of ubiquitous AI in FL-enabled 6G while considering scalable architecture as one of its key components. In another work \cite{wu2020cloud}, AI is used as a tool for establishing a link between cloud and edge computing platforms for IoT applications. This involves scalable solutions for distributing valuable resources that significantly reduce communication delay and storage requirements. Further, the authors in \cite{rahimi2021design} investigated edge computing in 5G considering the reliability, low latency, and scalability as the most important features for evaluating its functionality. This outcome can be comfortably extended to imparting scalable architecture for 6G-enabled IoT services.

The work in~\cite{wang2021federated} proposed a reconfigurable intelligent surface (RIS) to achieve quick model aggregation for over-the-air computation based FL. Here, with the deployment of RIS, the model ensures to provide higher accuracy compared to the state-of-the-art schemes. This scheme largely assists XAI models by ensuring robust interpretable solutions in the decision making phases. Very recently, the authors in~\cite{arisdakessian2022survey} surveyed the intrusion detection in the IoT ecosystem that encompasses the cybersecurity framework, XAI models and FL, for imparting robust protection of the IoT devices against various cyberattacks.

\subsubsection{Massive access} 
For the conventional bandwidth-limited technologies, incorporating massive access for IoT imparts a huge task considering the power and cost requirements. The 6G services incorporated with multi-antenna base stations could serve a large number of single-antenna IoT devices by deploying massive access features. The ever-increasing demand for high bandwidth efficiency, low latency, and management of heterogeneous data traffic in 6G services can be addressed using massive access techniques. In\cite{yuan2021noma}, the authors summarized the impact of 6G-based massive IoT by considering the multi-user channel coding, joint modulation, and linear spreading from the perspective of non-orthogonal multiple access technologies. The associated potential enabling technologies for 6G networks, its key performance indicators, and its dimensions concerning massive access techniques are summarized in\cite{akhtar2020shift}. 

The impact of massive access along with AI is widely found in digital twin applications, which incorporate the secure platform driven by blockchain and smart grid technologies \cite{lopez2021digital}. This involves the communication framework of $5$ GB, which can quite comfortably be extended for 6G-enabled IoT devices. In \cite{qi2020integration}, a joint beamforming design algorithm is developed for validating the energy harvesting capability of 6G-based IoT devices enabled with massive access capability. The impact of the diversity order in multi-user systems and finite block length theory plays a significant role in enabling massive access ability for IoT devices \cite{yuan2021noma}. A resource-hopping-based grant-free multiple access technique \cite{jang2021resource} in 6G IoT networks enables maximum device handling capability with better packet delay, latency, and interference, meeting the reliable requirements of IoT devices. The authors in \cite{lv2021big} implemented 6G-based big data analysis and targeted to meet large-scale IoT devices. This is enabled by using low-energy multiple access techniques that provide a higher success rate and the ability to analyze the data from multiple IoT devices. The grant-free non-orthogonal transmission schemes through multiple access are directed to enable 6G enabled IoT services \cite{abbas2020grant}. It is implemented using a DL-based joint user decoding and detection mechanism. This also incorporates a multi-layered model that drives multiple access capabilities for the system and triggers a massive contribution to IoT communication. With two novel algorithms, the massive access schemes are incorporated in \cite{mursia2020risma}, through which the sum rate performance is evaluated, and its outcome provides promising support for IoT applications enabled by 6G services. Moreover, in \cite{du2020machine}, spectrum management and channel estimation are addressed with the support of AI techniques for guaranteeing reliable low latency services for 6G-enabled IoT devices. Moreover, it ensures the provision of tight QoS on complex architectures and legacy networks.

\subsubsection{Green communication} 
As telecom vendors and operators are moving towards green communication services, supported by green networks, these use renewable energy, such as wind and solar energy, for their operations. Furthermore, the green networks incorporate power amplifiers for amplifying the generated power from renewable energy sources, which in turn supports the operation of the nodes in the network. Besides, the topology support of green networks is based on femtocells and agile bases. Green communication can be supported based on the cognitive radio services to optimize its operations and power requirements. The green communication services ensure power and computation-efficient communication services, which can be a perfect fit to operate IoT devices in conjunction with this environment-friendly technology. The ubiquitous intelligence in the current communication standards is driven through green communication services for provisioning better services, mobility, and resource management to a sustainable extent with optimized usage of renewable energy sources \cite{niu2020green,Yang2022Energy}. 

Hybrid whale spotted hyena optimization algorithm \cite{verma2020towards} outperforms the conventional energy-efficient data dissemination methods to provide supreme performance for green communication-enabled 6G-based massive IoT networks. Green communication can also be integrated with optical wireless communication technology, as reported by the authors in \cite{chowdhury2019role}. They have presented the effective means of free space optical communication facility considering the light fidelity for the faithful implementation in 6G-based IoT systems. Wang et al. \cite{wang2020base} developed a base station wakeup strategy, which could be applied to 6G services that consider the energy states by switching between on-grid and renewable services for the targeted IoT applications. The Transfer Asynchronous Advantage Actor-Critic algorithm \cite{zhang2021learning} helps to analyze the resource allocation issue considering the energy requirements as one of the optimal resources for a communication system. This flexible scheme outperforms other ML approaches to provide learning-based optimized green communication services. In a vehicle-to-network scenario, the UAV-aided system is enabled to establish an architecture that provides a green communication link between the UAV, satellite, and UAV and IoT networks \cite{dai2020uav}. 

\subsection{Summary}
In this section, we highlighted the role of edge XAI structure and 6G services, along with their enabling technologies, platforms, characteristics, and challenges, to deliver trustworthy IoT applications. The great success of XAI frameworks in support of IoT applications is backed by distributed services offered through RL, FL, and blockchain-based services. Deploying XAI structures on edge devices for IoT applications, such as healthcare and smart cities, draws the attention of end-users due to their interpretable characteristics and quick response time as well as better accessibility. Nevertheless, edge XAI structures will be a great means for the long-term and massive integration of large numbers of IoT devices operating on a large volume of data. IoT components need to utilize the edge XAI infrastructure for faithful integration of data analytics engines and cloud services. Currently, the integration of 6G services into IoT applications is limited due to the lack of enabling services, which are based on the deployment of supporting technologies. While huge demand exists for mURLLC, scalable architecture, and massive access to sensitive real-time communications, the green communication concept over IoT applications could support the modern communication area to a larger extent.  

\section{Future Research Directions and Consideration of XAI for IoT}
\label{sec:future}
This survey discussed the key features of XAI, focusing on the design of IoT systems, considering the XAI enabling technologies, and emphasizing the transparency of the decision-making process in IoT applications. However, there is still a huge scope for the successful implementation of XAI in practical IoT systems, including THz and semantic communication, high energy-efficiency issues, synchronization, a new explainable model for signal processing perspectives, such as joint channel detection and transmission with XAI for IoT networks, and low cost and low power consumption hardware design for XAI devices over IoT. Some discussions on these issues are provided as follows for future researchers to grasp promising solutions.

\subsection{THz Communication}
Revolutionary enhancement of data rate is witnessed in current wireless communication systems. On the other hand,  ubiquitous devices are foreseen due to the immersion of the IoT paradigm. Thus, it is expected to reach up to 1 Tbps peak data rates and 10 Gbps experienced data rates for users for 6G. To meet the requirements, the Terahertz technology is considered one of the enabling technologies owing to the abundant frequency band. The wireless channel at THz frequency is highly uncertain and dynamic THz technology is considered one of the enabling technologies of tomorrow's 6G wireless systems. Due to the abundant frequency band and the high susceptibility to blockage and molecular absorption, THz frequencies can potentially provide significant wireless capacity performance gains and enable high-resolution environment sensing. It can be considered as a means to provide communication and sensing services for IoT applications \cite{tong2022environment, Tong2021Joint}. However, the challenges of modeling, network design, and the emergence of optimization require us to rethink the conventional physical (PHY) layer and networking procedures. The large volume of data generated in the IoT devices can be analyzed using AI algorithms to help improve the performance of the THz communication, which in turn guarantees the QoS of IoT applications.

\subsection{Semantic Communication}
With the objective of extracting insights from the transmitted information through the matched knowledge base, we need to design effective methods to improve semantic communication in the IoT scenario \cite{gunduz2022beyond,xu2022edge}. For the next-generation communication among IoT devices and the infrastructure, an adaptive and intelligent communication system is about to move forward. Semantic communication can improve the secondary market in an adaptive manner \cite{tong2021federated,tong2021federated2}, where the users with XAI capacity at the edge devices can dynamically detect the decision made on the interpretation of semantic information. DL-based semantic communication is one of the key techniques to improve transmission efficiency. However, the choice of XAI models to balance transparency in decision-making and efficient transmission is a very important problem in semantic communication for IoT. The transparency and trustworthiness of IoT frameworks can bring new challenges to effective semantic communication. 

\subsection{High Energy Efficiency Issues}
In recent years, many new techniques have been proposed for the design of energy-efficient IoT devices and their associated edge devices. However, the ubiquitous nature of IoT makes itself responsible for energy drainage. Understanding and application of key energy-efficient techniques may vary under different scenarios and types of IoT frameworks, that in turn make the problems more complex. In AI-enabled IoT solutions, large-scale network deployment has become an inevitable trend, and the combination of XAI techniques can effectively solve this problem. For instance, an IoT framework with allocated resources for the network and devices can provide maximum efficiency and lower energy consumption. However, the problem is still challenging to solve due to the distributed nature of IoT settings and the models adaptable to the scenario. In other words, designing a highly flexible and low-complexity XAI model for further enhancing the trustworthiness in an energy-efficient manner for IoT systems is an important research direction.

\subsection{Hardware Design for XAI Devices over IoT}
The large volume of data produced by the IoT devices is processed locally rather than offloading to the cloud servers, which can be supported by distributed XAI technologies. Such technological changes require significant strides in the energy efficiency of IoT devices, which have become a challenge for hardware design. To implement XAI in IoT devices, the corresponding protocols should be renewed. For example, the synchronization protocol of the distributed devices should be designed when there exists some information exchange among XAI-enabled devices. Neural networks are predominantly used in AI systems. The numbers of neurons, neural layers, and activation functions influence the number of operations and the computational complexity. Associated hardware to run neural networks should be designed to save energy, which is the key to the implementation of this model.   

\subsection{Distributed XAI}
Undoubtedly, the management of a distributed range of XAI systems to draw faithful and trustworthy conclusions on the data collected from IoT devices is one of the most challenging issues in the deployment of transparent IoT applications driven by XAI. Nowadays, to improve the delay encountered in the integration of the decisions made from the distributed entities of XAI models, many high computing devices, such as GPUs, are available. However, imparting the transparency features in their decision-making phases and combining the decisions made from the distributed XAI platform is a challenging issue. It is, therefore, necessary to gather evidence and adopt alternate solutions in distributing IoT data to the XAI models and accumulating the decisions made by the models for trustworthy interpretation of stage-wise decisions made by those models. 

\subsection{XAI Security and Privacy}
Beyond the aforementioned technical challenges, security and privacy are major issues to be considered while collecting data from IoT devices and using them to train the XAI models. The study of defense mechanisms against the attack in XAI systems in terms of imparting privacy and confidentiality properties for the explainable methods still needs to be explored by the research community. Moreover, investigation of robustness in the security of XAI with respect to different neural network architectures is a demand among the use cases. Moreover, security challenges remain prevalent when the XAI models operate on the data generated from IoT devices. From the perspective of the cybersecurity domain, the threat models for XAI methods need to be expanded to reflect real-world settings and secure the data gathered by IoT devices. Furthermore, consistency in the design prospects of mitigating black-box attacks is required to study and analyze the correctness and confidence of XAI methods. 

\subsection{Accountable and Systematic Decision Making}
Quality of decision-making is largely improved with XAI, and it ensures the stakeholders take up responsibility with confidence. Based on the domain and applications, XAI models assist in providing better system requirements for design, measurements, and periodic testing by the users. They offer exemplary opportunities for automated decision-making and guarantee to provide specific, accountable, and systematic ways of using the core values and principles of the human decision-making process. However, to gain maximum benefit from deploying particular tools for automated and systematic decision making, even before they are completely understood, research on developing robust XAI paradigms and their deployment and testing over crucial use cases are anticipated. 

\section{Conclusion}
\label{sec:conclusion}

The explosive boom of IoT and the rising popularity of automation and remote monitoring and control over the Internet have created concerns over the trust of end-users. This is due to an inefficient architecture that has less concern over the transparency of operations in the network of smart devices. Researchers and industry experts are embracing the use of XAI as a promising technology to help overcome these challenges. XAI frameworks are capable of implementing new capabilities and solutions towards enabling trustworthiness in IoT devices and networks. However, due to resource constraints in IoT devices, it is challenging to implement full-fledged trustworthy services for the end-users. 

In this paper, we presented a comprehensive survey on XAI solutions for IoT systems. We started with an overview of XAI characteristics and their salient features, followed by a discussion of its role in modern use cases. A description of the demand for developing trustworthy systems, emphasizing IoT applications and recent standards, was provided. We further provided the state-of-the-art past and progressive research activities in the field of XAI and IoT regarding the considerations towards the development of explainable systems. We also highlighted IoT use cases based on the explainable features imparted in such systems. Furthermore, we extensively explored XAI over IoT adaptive solutions using emerging architectures based on 5G services, cloud services, and big data management. Moreover, we extended the explainable capability of IoT frameworks to newer domains, such as IIoT and other smart applications. Based on the survey regarding the XAI over IoT, we discussed the research challenges that require immediate attention from the research community. We believe that this paper outlines the research gaps in the field of XAI and its impact on IoT and will assist the readers in perceiving an overview of the recent progress in this domain and drive towards future works.
\ifCLASSOPTIONcaptionsoff
  \newpage
\fi
\bibliographystyle{IEEEtran}
\bibliography{Ref.bib}
\begin{IEEEbiography}
[{\includegraphics[width=1in,height=1.25in,clip,keepaspectratio]{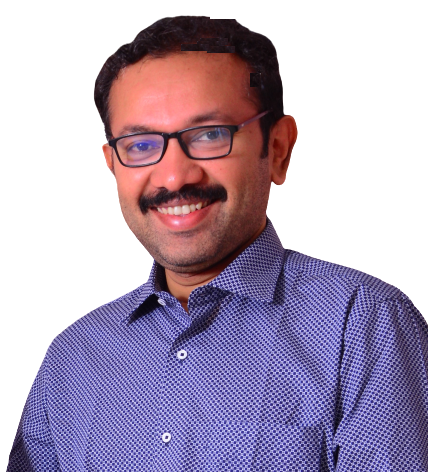}}]
{Senthil Kumar Jagatheesaperumal}
 received his B.E. degree from Madurai Kamaraj University, Tamilnadu, India in 2003. He received his Post Graduation degree from Anna University, Chennai, in 2005. He received his Ph.D. degree in Information \& Communication Engineering from Anna University, Chennai in 2017. He is currently working as an Associate Professor in the Department of Electronics and Communication Engineering, Mepco Schlenk Engineering College, Sivakasi, Tamilnadu, India. He received two funded research projects from National Instruments, USA each worth USD 50,000 during the years 2015 and 2016. He also received another funded research project from IITM-RUTAG during 2017 worth Rs.3.97 Lakhs. His area of research includes Robotics, the Internet of Things, Embedded Systems, and Wireless Communication. During his career, he has published various papers in International Journals and conferences. He is a Life Member of IETE and ISTE.
\end{IEEEbiography}
\begin{IEEEbiography}
[{\includegraphics[width=1in,height=1.25in,clip,keepaspectratio]{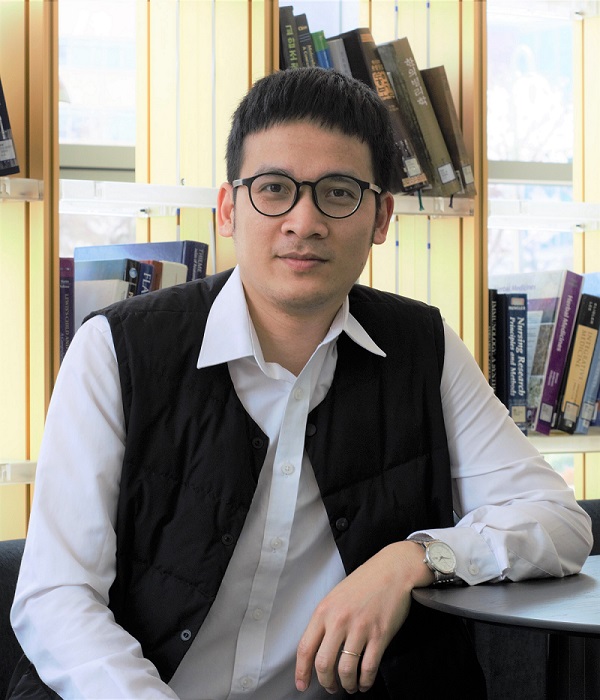}}]
{Quoc-Viet Pham} (Member, IEEE) received the B.S. degree in electronics and telecommunications engineering from the Hanoi University of Science and Technology, Vietnam, in 2013, and the Ph.D. degree in telecommunications engineering from Inje University, Gimhae, Republic of Korea, in 2017. He has been a Research Professor with Pusan National University, Republic of Korea, since Jan. 2020.

He is specialized in applying convex optimization, game theory, and machine learning to analyze and optimize edge computing and future wireless communications. He has been granted the Korea NRF Funding for outstanding young researchers for the term 2019–2024. He was a recipient of the Best Ph.D. Dissertation Award from Inje University in 2017, the Top Reviewer Award from the IEEE Transactions on Vehicular Technology in 2020, and the Golden Globe Award from the Ministry of Science and Technology (Vietnam) in 2021. He is also an editor of Journal of Network and Computer Applications (Elsevier), Scientific Reports (Nature), and Frontiers in Communications and Networks, and was a lead guest editor of the IEEE Internet of Things Journal.
\end{IEEEbiography}
\begin{IEEEbiography}
[{\includegraphics[width=1in,height=1.25in,clip,keepaspectratio]{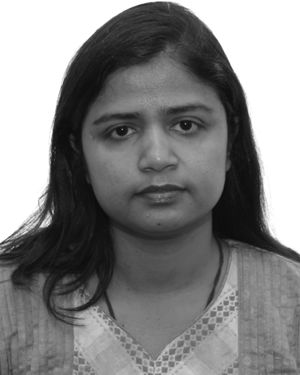}}]
{Rukhsana Ruby} (Member, IEEE) received her Masters degree from University of Victoria, Canada on 2009, and PhD degree from The University of British
Columbia, Canada on 2015. From the broader aspect, her resource interests include the management and optimization of next generation wireless networks. She has authored nearly 75 technical papers of well-recognized journals and conferences. She is the recipient of several awards or honors, notable among which are Canadian NSERC Postdoctoral Fellowship, IEEE Exemplary Certificate (IEEE Communications Letters, 2018 and IEEE Wireless Communications Letters,
2018) and Outstanding Reviewer Certificate (Elsevier Computer Communications, 2017). She has served as the lead guest editor for the special issue on
NOMA techniques under EURASIP JWCN on 2017, and currently is serving as an associate editor to this journal. Besides, she has also been serving as a
technical program committee member in various reputed conferences.
\end{IEEEbiography}
\begin{IEEEbiography}
[{\includegraphics[width=1in,height=1.25in,clip,keepaspectratio]{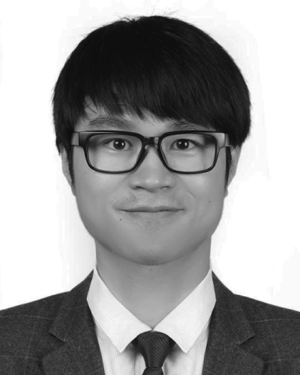}}]
{Zhaohui Yang} (Member, IEEE) received the B.S. degree in information science and engineering from the Chien-Shiung Wu Honors College, Southeast University, Nanjing, China, in June 2014, and the Ph.D. degree in communication and information system from the National Mobile Communications Research Laboratory, Southeast University, in May 2018. He is currently a Post-Doctoral Research Associate with the Center for Telecommunications Research, Department of Informatics, King’s College London, U.K., where he has been involved in the EPSRC SENSE Project. He has guest edited a 2020 Feature topic of the IEEE Communications Magazine on Communication Technologies for Efficient Edge Learning. His research interests include federated learning, reconfigurable intelligent surface, UAV, MEC, energy harvesting, and NOMA. He was a TPC member of the IEEE ICC from 2015 to 2020 and Globecom from 2017 to 2020. He was an Exemplary Reviewer of the IEEE Transactions on Communications in 2019. He serves as the Co-Chair for the IEEE International Conference on Communications (ICC), 1st Workshop on Edge Machine Learning for 5G Mobile Networks and Beyond, in June 2020, the IEEE International Symposium on Personal, Indoor, and Mobile Radio Communication (PIMRC) Workshop on Rate-Splitting, and Robust Interference Management for Beyond 5G, August 2020.
\end{IEEEbiography}
\begin{IEEEbiography}
[{\includegraphics[width=1in,height=1.25in,clip,keepaspectratio]{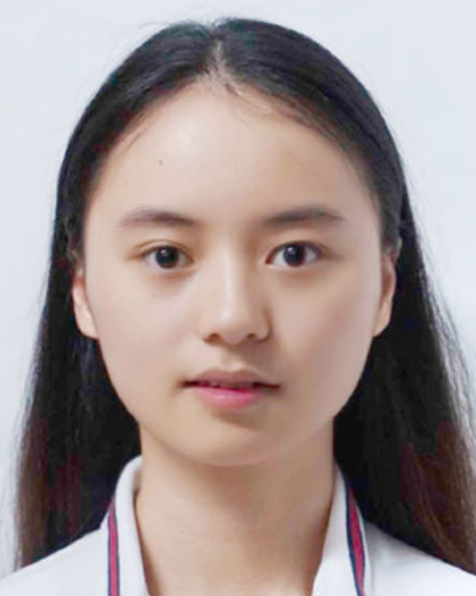}}]
{Chunmei Xu} (Student Member, IEEE) received the B.Eng. degree in information engineering in 2017, from the School of Information Science and Engineering, Southeast University, Nanjing, China, where she is currently working toward the Ph.D. degree in information and communication engineering. Her current research focuses on intelligent wireless communications.
\end{IEEEbiography}
\begin{IEEEbiography}
[{\includegraphics[width=1in,height=1.25in,clip,keepaspectratio]{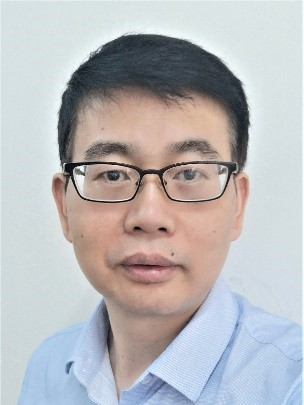}}]
{Zhaoyang Zhang}  (Senior Member, IEEE) received the Ph.D. degree from Zhejiang University, Hangzhou, China, in 1998.,He is currently a Qiushi Distinguished Professor at Zhejiang University. He has coauthored more than 300 peer-reviewed international journals and conference papers, including eight conference best papers awarded by IEEE ICC 2019 and IEEE GlobeCom 2020. His research interests include fundamental aspects of wireless communications and networking, such as information theory and coding theory, network signal processing and distributed learning, AI-empowered communications and networking, and synergetic sensing, computing, and communication. He was awarded the National Natural Science Fund for Distinguished Young Scholars by NSFC in 2017. He is serving or has served as an Editor for IEEE Transactions on Wireless Communications, IEEE Transactions on Communications, and IET Communications, and the General Chair, the TPC Co-Chair, or the Symposium Co-Chair for PIMRC 2021 Workshop on Native AI Empowered Wireless Networks, VTC-Spring 2017 Workshop on HMWC, WCSP 2013/2018, Globecom 2014 Wireless Communications Symposium. He was also a Keynote Speaker for APCC 2018 and VTC-Fall 2017 Workshop NOMA.
\end{IEEEbiography}

\end{document}